%% file: elsarticle-template-num.tex
\PassOptionsToPackage{authoryear}{natbib}
\documentclass[preprint, 12pt]{elsarticle}

\UseRawInputEncoding
\usepackage[utf8]{inputenc}
\usepackage{enumitem}
\usepackage{amsmath,amsfonts}
\usepackage{array}
\usepackage[left]{lineno} 
\usepackage{textcomp}
\usepackage{stfloats}
\usepackage{url}
\usepackage{verbatim}
\usepackage{graphicx}

\usepackage{amssymb}
\usepackage{booktabs}
\usepackage{bbm}
\usepackage{multirow}
\usepackage{siunitx}
\usepackage{caption}
\usepackage{subcaption}
\usepackage{algorithm}
\usepackage{algpseudocode}

\DeclareMathOperator{\diag}{diag}
\DeclareMathOperator{\Bernoulli}{Bernoulli}

\usepackage{amssymb}
\usepackage{amsmath}

\begin{document}

\begin{frontmatter}



\title{Uncertainty-Aware Deep Learning for Wildfire Danger Forecasting}


\author[NOA,UVEG]{Spyros Kondylatos\corref{cor1}} 
\author[NOA]{Nikolas Papadopoulos} 
\author[UVEG]{Gustau Camps-Valls}
\author[NOA,ARCH]{Ioannis Papoutsis}

\cortext[cor1]{Corresponding author: skondylatos@noa.gr}

\affiliation[NOA]{organization={Orion Lab, National Observatory of Athens \& National Technical University of Athens},
            addressline={Iroon Polytechneiou 9, Zografou}, 
            city={Athens},
            postcode={15772}, 
            country={Greece}}
            
\affiliation[UVEG]{organization={Image Processing Laboratory, Universitat de València},
            addressline={C/ Cat. Agustín Escardino Benlloch, 9, Paterna}, 
            city={Valencia},
            postcode={46980}, 
            country={Spain}}

\affiliation[ARCH]{organization={Archimedes, Athena Research Center},
            addressline={Artemidos 1, Maroussi}, 
            city={Athens},
            postcode={15125}, 
            country={Greece}}

\begin{abstract}

Wildfires are among the most severe natural hazards, posing a significant threat to both humans and natural ecosystems.
The growing risk of wildfires increases the demand for forecasting models that are not only accurate but also reliable.
Deep Learning (DL) has shown promise in predicting wildfire danger; however, its adoption is hindered by concerns over the reliability of its predictions, some of which stem from the lack of uncertainty quantification.
To address this challenge, we present an uncertainty-aware DL framework that jointly captures epistemic (model) and aleatoric (data) uncertainty to enhance short-term wildfire danger forecasting.
In the next-day forecasting, our best-performing model improves the F1 Score by 2.3\% and reduces the Expected Calibration Error by 2.1\% compared to a deterministic baseline, enhancing both predictive skill and calibration.
Our experiments confirm the reliability of the uncertainty estimates and illustrate their practical utility for decision support, including the identification of uncertainty thresholds for rejecting low-confidence predictions and the generation of well-calibrated wildfire danger maps with accompanying uncertainty layers.
Extending the forecast horizon up to ten days, we observe that aleatoric uncertainty increases with time, showing greater variability in environmental conditions, while epistemic uncertainty remains stable. 
Finally, we show that although the two uncertainty types may be redundant in low-uncertainty cases, they provide complementary insights under more challenging conditions, underscoring the value of their joint modeling for robust wildfire danger prediction.
In summary, our approach significantly improves the accuracy and reliability of wildfire danger forecasting, advancing the development of trustworthy wildfire DL systems.
\end{abstract}








\end{frontmatter}



\section{Introduction}
\label{sec:intro}

Natural hazards such as floods, storms, droughts, heatwaves, and wildfires pose significant threats to societies and ecosystems around the world \citep{guikema_natural_hazards}.
Their frequency and intensity are being amplified by climate change and anthropogenic activities \citep{KALANTARI2019393}, resulting in widespread disruptions to human livelihoods and causing long-lasting damage to natural ecosystems.
As these risks intensify, the development of robust and reliable systems for modeling and understanding natural hazards becomes critical to support disaster management efforts \citep{camps-valls_artificial_2025}.
This need is timely, as authorities are compelled to reconsider traditional risk mitigation policies to better address the challenges of a rapidly changing climate \citep{moreira_wildfire_2020, sun_applications_2020}.

In this evolving risk landscape, Artificial Intelligence (AI) has emerged as a transformative tool for supporting natural hazard applications, from detection and monitoring to forecasting and response \citep{HIMEUR202244}.
In particular, Deep Learning (DL) has shown significant promise for modeling the complex dynamics of Earth systems \citep{reichstein_deep_2019}, leveraging the growing availability of diverse data sources from Earth Observation (EO) \citep{soille_versatile_2018} and climate information derived from reanalysis and observations \citep{salcedo_analysis}.
Despite these advances, the widespread integration of DL in disaster management contexts remains limited.
Barriers to adoption include concerns about the transparency, reliability, and calibration of model outputs \citep{persello_2021}, with DL models frequently producing overconfident predictions \citep{srivastava_dropout}.
Addressing these challenges is essential for ensuring that DL-based tools can be responsibly embedded into natural hazard management systems \citep{bostrom_trust, albahri_systematic_2024}. 
A key step in this direction is the quantification of uncertainty in model predictions, which can serve as a foundational approach for improving the reliability of DL models \citep{gawlikowski_survey_2021, tuia_toward_2021}.

Among natural hazards, wildfires represent a particularly compelling case for the integration of uncertainty-aware modeling.
Wildfires play a vital ecological role by influencing the global carbon cycle and shaping ecological development through disturbance and regeneration \citep{pausas_burning_2009}; yet, they pose severe threats to human lives, infrastructure, and the economy \citep{pettinari_fire_2020}. 
Their frequency and intensity are increasingly influenced by humans and climate change, which disrupt natural fire regimes \citep{pausas_wildfires_2021}.
Moreover, wildfires are inherently stochastic phenomena governed by complex interactions among biophysical drivers operating across multiple spatial and temporal scales \citep{archibald_defining}.
These underlying complexities, combined with the ongoing environmental shifts resulting from climate change, introduce significant uncertainty in modeling, particularly for models trained on historical observations.

Short-term wildfire danger forecasting plays a critical role in wildfire preparedness, guiding timely interventions and effective resource allocation. 
In this context, the reliability of predictive models is essential for building trust toward forecast-driven decisions. 
Recent advances have shown that DL can outperform traditional indices, such as the Canadian Fire Weather Index, in predicting wildfire danger \citep{kondylatos_wildfire_2022}. 
However, despite these performance gains, current DL-based wildfire forecasting studies have largely overlooked the importance of model uncertainty.
As the application of DL to wildfire danger forecasting continues to grow, investigating the uncertainty and calibration of these models is essential to enhance their reliability and trust.

In this work, we use uncertainty-aware DL to forecast short-term wildfire danger.
Our area of interest is the Mediterranean basin, an area particularly vulnerable to wildfires and projected to face increasing threats shortly \citep{ruffault_increased_2020}.
Our approach treats wildfire danger forecasting as a supervised classification task, where both epistemic and aleatoric uncertainty are relevant \citep{hullermeier_aleatoric_2021}.
We employ a unified DL framework that estimates both epistemic and aleatoric uncertainty, enabling a comprehensive characterization of predictive uncertainty.
Epistemic uncertainty is captured using Bayesian Neural Networks (BNNs) \citep{jospin_hands-bayesian_2022} and Deep Ensembles (DEs) \citep{lakshminarayanan_simple_2017}.
Aleatoric uncertainty is estimated through modeling a distribution over the network logits, which captures the inherent noise in the labels \citep{collier_simple_2020}.
This integrated approach allows us to examine the combined effects of epistemic and aleatoric components on predictive performance and uncertainty reliability and to assess whether they offer complementary information or exhibit redundancy.

The key contributions and insights of this work are summarized as follows:
\begin{itemize}
    \item We develop an uncertainty-aware DL framework that jointly models epistemic and aleatoric uncertainty for predicting the next day's wildfire danger. Through comparative analysis with models that account for only one or neither type of uncertainty, we demonstrate improvements in predictive performance, calibration, and reliability of uncertainty estimates.
    \item We extend our analysis across multiple forecasting temporal horizons, ranging from one to ten days ahead, to investigate the temporal relationship between uncertainty and model performance. Doing so, we reveal that aleatoric uncertainty increases with longer forecast horizons---reflecting the accumulation of stochasticity in environmental inputs---while epistemic uncertainty remains relatively stable across time.
    \item We demonstrate the practical utility of uncertainty-aware modeling for supporting decision-making in wildfire danger forecasting by i) generating wildfire danger maps accompanied by disentangled uncertainty estimates, ii) providing practical ways for rejecting less accurate predictions based on uncertainty thresholds; and iii) demonstrating improved model calibration, thus mitigating overconfidence in DL model outputs.
    \item By disentangling epistemic and aleatoric uncertainty, we show that, although their information is largely redundant in low total uncertainty (less challenging) samples, they provide complementary insights in high total uncertainty (more challenging) cases.
\end{itemize}

\section{Related Work} 

\subsection{Uncertainty in Deep Learning}
In ML, uncertainty is typically classified into two main types: epistemic (model) uncertainty and aleatoric (data) uncertainty. 

\noindent\textit{Epistemic uncertainty} arises from a model’s lack of knowledge and can be reduced with more data or model improvement.
This uncertainty is usually captured using Deep Ensembles (DEs) \citep{lakshminarayanan_simple_2017} and Bayesian Neural Networks (BNNs) \citep{jospin_hands-bayesian_2022}.
DEs employ multiple independently trained deterministic NNs, using the variance of their predictions as estimates of uncertainty.
BNNs incorporate prior distributions into network parameters and estimate posteriors using Bayesian methods like Markov Chain Monte Carlo \citep{bishop_pattern_2009} or Variational Inference (VI) \citep{barber_ensemble_nodate}. 
VI-based methods \citep{blundell_weight_2015, hernandez-lobato_probabilistic_2015, gal_dropout_2016}, along with normalizing flows \citep{louizos_multiplicative_2017}, have become more popular in the DL era due to their computational efficiency and scalability for uncertainty estimation in large-scale DL models.

\noindent\textit{Aleatoric uncertainty} stems from inherent noise in the data and cannot be reduced, even with additional data samples.
To estimate this uncertainty, post hoc methods have been applied to pretrained deterministic NNs without modifying the original model architecture \citep{ramalho_density_2019, oberdiek_classification_2018, lee_gradients_2020}.
Moreover, test-time data augmentation methods do inference over multiple augmented versions of the input, with the variability among these outputs serving as a proxy for uncertainty \citep{ayhan_test-time_2018, wang_aleatoric_2019}.
Another class of approaches incorporates architectural modifications to enable the model to learn and predict uncertainty directly during training, typically by parameterizing a probability distribution over the outputs during training \citep{malinin_predictive_2018, sensoy_evidential_2018, nandy_towards_2021}.
Notably, \citet{kendall_what_2017} proposed a method within this category, introducing a heteroscedastic uncertainty model that learns to predict input-dependent noise by placing a Gaussian distribution over the logits of a softmax classifier.

\subsection{Uncertainty-Aware Deep Learning for Natural Hazards}

In seismology, \citet{mousavi_bayesian-deep-learning_2020} employed BNNs to estimate earthquake locations, modeling prediction errors concerning uncertainties in epicentral distances and travel times. 
Similarly, \citet{bueno_volcano-seismic_2020} applied probabilistic Bayesian DL to improve volcano-seismic monitoring, linking uncertainty estimates to stages of volcanic unrest.
Moreover, MC Dropout has been utilized for seismic event detection \citep{gamboa-chacon_analysis_2025}. 
In the meteorological domain, \citet{wang_deep_2019} integrated DL-based weather forecasting with uncertainty quantification and DEs, achieving enhanced accuracy and reliability over numerical weather predictions. 
For drought forecasting, \citet{ferchichi_evidential_2025} introduced an evidential DL approach using Dirichlet distributions to interpret both aleatoric and epistemic uncertainties as evidence associated with model outputs.
\citet{klotz_uncertainty_2022} proposed a benchmark for uncertainty estimation in hydrological prediction using Mixture Density Networks and MC Dropout. 
In landslide modeling, \citet{wang_quantification_2023} applied MC simulation techniques to quantify model uncertainty in landslide displacement forecasting.
\citet{kondylatos_probabilistic_2025} provided a pipeline for aleatoric uncertainty estimation by modeling heteroscedastic label noise in several high-stakes EO applications.
Finally, \citet{zhang_calibration_2025} proposed methods for improving calibration and providing entropy-based uncertainty quantification in DL models for drought detection.

\subsection{Deep Learning in wildfire-related applications}

DL has seen increasing adoption across various wildfire-related applications, including wildfire danger forecasting, active fire detection, burned area mapping, and fire spread prediction \citep{jain_review_2020, ghali_2023}.
Convolutional Neural Networks (CNNs) have shown strong performance in detecting active fires from satellite and aerial thermal imagery \citep{bouguettaya2022review, ghali2022deep}, while transformer-based models outperform spatial and temporal-aware models in segmenting active fire pixels from VIIRS satellite data \citep{zhao_tokenized_2023}.
For post-fire analysis, DL models often surpass traditional indices (e.g., Normalized Burn Ratio) in mapping burned areas from high-resolution imagery like Sentinel-2 \citep{knopp2020deep, sdraka2024floga}.
For wildfire spread prediction, convolutional models have proven effective in capturing spatial \citep{shadrin2024wildfire} or spatiotemporal \citep{burge_convolutional_2021} patterns. 
Models like FireCast \citep{radke_firecast_2019} employ CNNs on historical fire data to identify high-risk spread zones. 
\citet{hodges_wildland_2019} introduced a Deep Convolutional Inverse Graphics Network for forecasting spread up to six hours ahead. 
Large-scale datasets such as Next Day Wildfire Spread \citep{huot2022next} and WildfireSpreadTS \citep{NEURIPS2023_ebd54517} have enabled daily prediction and benchmarking of DL approaches.

\textbf{Wildfire danger forecasting.} Particularly, for wildfire danger forecasting, DL methods have been applied, offering promising results across various regions and timescales. 
\citet{zhang_forest_2019} employed a CNN to predict next-day wildfire danger in China’s Yunnan province and later extended their work to generate seasonal global fire susceptibility maps using CNNs and Multilayer Perceptrons \citep{zhang_deep_2021}.
\citet{Prapas_2023_ICCV} predicted global wildfire danger using a teleconnection-driven vision transformer capable of treating the Earth as one interconnected system.
Similarly, \citet{bergado_predicting_2021} used fully convolutional networks to produce daily wildfire probability maps for the upcoming week. 
In Chile, \citet{bjanes_deep_2021} developed an ensemble of DL models for wildfire susceptibility mapping. 
\citet{huot_deep_2020} approached wildfire forecasting as an image segmentation task, using CNNs to predict danger levels in the USA at multiple temporal resolutions. 
\citet{kondylatos_wildfire_2022} applied temporal-aware networks for next-day fire danger prediction in the Eastern Mediterranean, integrating explainable AI techniques to interpret the model’s outputs.

\section{Data \& Motivation}

\subsection{Dataset \& Task Formulation }
\label{sec:dataset}

The dataset used for predicting short-term wildfire danger is extracted from Mesogeos \citep{mesogeos_kondylatos}, a data cube with a daily temporal resolution and a spatial resolution of $1 km \times 1 km$, covering the Mediterranean region from 2006 to 2022. 
Mesogeos integrates a wide array of variables associated with fire danger.
These include satellite-derived vegetation indices (Normalized Difference Vegetation Index and Leaf Area Index), day/night land surface temperatures, land cover classes, soil moisture, meteorological observations (temperature, wind speed, wind direction, dewpoint temperature, surface pressure, relative humidity, precipitation, surface solar radiation), geomorphological characteristics (elevation, slope, aspect, and curvature), and human activity indicators (distance from roads, population). 
These variables collectively capture key drivers of wildfire risk: weather conditions, vegetation, and topography affect the rate of fuel drying and fire spread, while human activity shapes ignition likelihood and fire regimes.
Additionally, Mesogeos contains wildfire ignition points, associated burned areas, and fire sizes for 25,722 fire events throughout the study period, providing essential ground-truth information for model training and evaluation.

The task formulation and dataset structure follow the methodology outlined in Track A of the original Mesogeos study. 
We summarize the key aspects here and refer readers to the original work for further details.
For each cell, wildfire danger on the day $t$ is defined as the probability of a wildfire occurrence based on the preceding conditions of fire driver variables.
For each target pixel, we extract a time series of dynamic (time-variant) variables observed over the 55 days preceding the prediction day $t$ (i.e., days $t - 1, t - 2, \ldots, t - 55$).
We also include static (time-invariant) features, which are repeated across the temporal dimension to match the shape of the dynamic input.
This study predicts fire danger in various temporal horizons.
For each forecast lead time, a 45-day window is used from the extracted time series to serve as input for training a DL model targeting that specific horizon.
Further details are provided in Sec.~\ref{sec:experiments}.

The prediction task is structured as a binary classification problem, where the positive class corresponds to increased danger, while the negative class represents low-danger cases. 
High-danger samples correspond to the ignition points of all wildfires in the dataset, with the associated burned area size serving as an indicator of the corresponding danger level. 
Low-danger samples are randomly drawn from locations at least 62 km away from any known ignition point.
The dataset is well-balanced, as the number of negative samples is twice that of the positives, with negative samples stratified by land cover type to ensure consistency with the distribution of the positive class.
The years 2006-2019 are used for training, 2020 is used as the validation set, and 2021-2022 are used for testing.

\begin{figure*}
    \centering
    \includegraphics[width=1.0\linewidth]{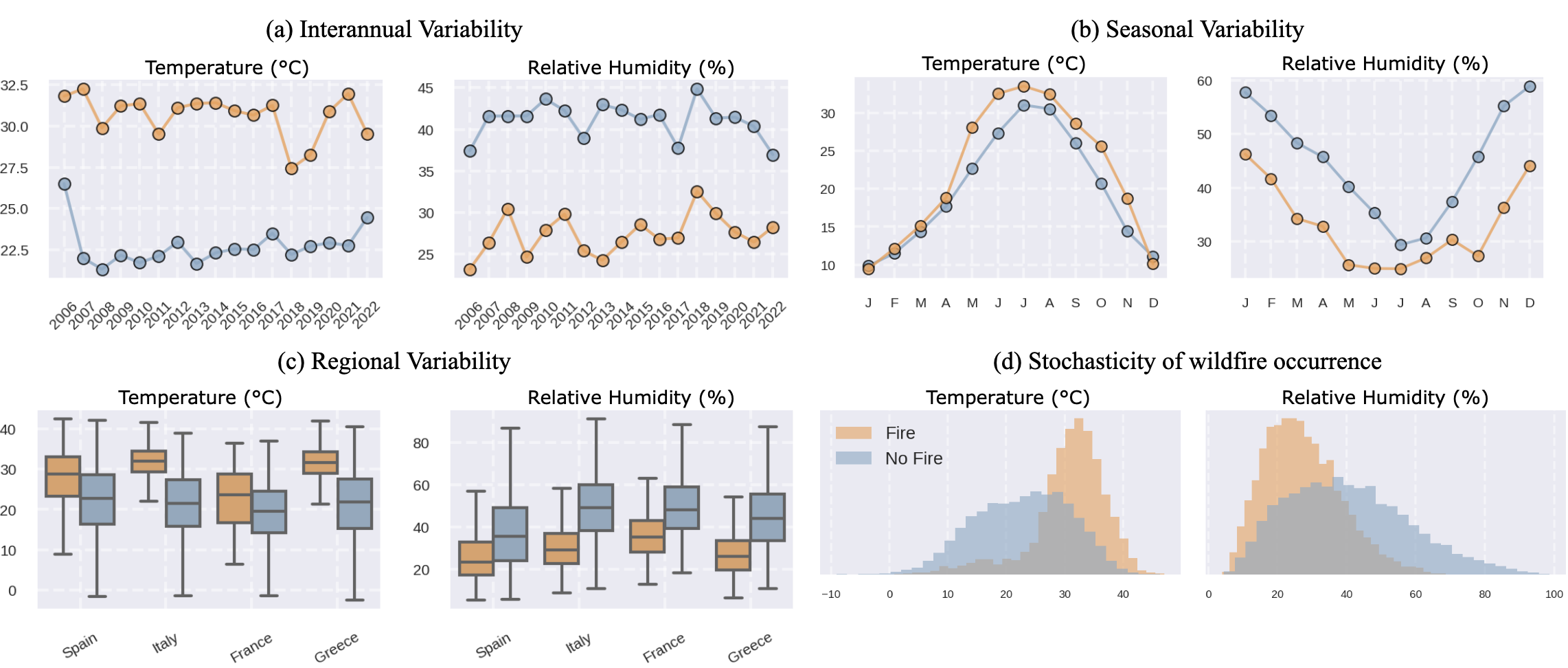}
    \caption{Distributional shifts in temperature and relative humidity for fire and non-fire pixels in the dataset. (a) Interannual variability: average values across different years (including training, validation, and test sets); (b) Seasonal variability: monthly averages of the values; (c) Regional variability: box plots of the values for four Mediterranean countries—Spain, Italy, France, and Greece; (d) Joint density plots showing the overlap in temperature and relative humidity distributions between fire and non-fire pixels.}
    \label{fig:distribution_shift}
\end{figure*}

\subsection{The need for uncertainty in wildfire danger forecasting} 
\label{sec:uncertainty_need}

Wildfire danger forecasting is a complex task that is inherently uncertain across multiple dimensions.
Fire ignition itself is a stochastic process arising from intricate and often unpredictable interactions between environmental variables and anthropogenic influences \citep{elia_modeling_2019, deste_modeling_2020}.
Predicting fire danger requires modeling a range of dynamic and heterogeneous inputs such as meteorological conditions, vegetation indices, and fuel moisture indicators, which are non-stationary, noisy, and subject to distributional shifts \citep{hantson_2016}.
All these factors induce both epistemic and aleatoric uncertainty in modeling, posing challenges to the generalization capacity of data-driven models.

One source of epistemic uncertainty stems from the limited availability of fire data, with fire events being exceedingly rare (e.g., a fire-to-non-fire ratio of $5 \times 10^{-7}$ in Mesogeos).
However, this actual distribution is difficult to preserve in DL classification datasets.
In practice, a more balanced dataset is constructed to enable learning, which introduces a mismatch between the training and actual distributions.
Additional epistemic uncertainty arises from distributional shifts in fire-driving variables across spatial and temporal dimensions.
Figure \ref{fig:distribution_shift} illustrates these shifts in the extracted dataset for temperature and relative humidity, two key variables for predicting fire danger.
Particularly, Figures \ref{fig:distribution_shift}-(a) and \ref{fig:distribution_shift}-(b) highlight their seasonal and interannual variability for fire and non-fire pixels, revealing differences across yearly (spanning training, validation, and test sets) and monthly averages. 
Figure \ref{fig:distribution_shift}-(c) further demonstrates regional shifts by comparing distributions across four representative Mediterranean countries: Spain, Italy, France, and Greece.
Apart from problem-specific sources, epistemic uncertainty is also tied to the difficulty of fully constraining model parameters during training.

Aleatoric uncertainty is primarily driven by the inherently stochastic nature of wildfire occurrence, where similar environmental and meteorological conditions may or may not result in fire ignition \citep{prapas_deep_2021}.
Figure \ref{fig:distribution_shift}-(d) shows that, although temperature and relative humidity can often differentiate fire from non-fire pixels, their distributions still exhibit substantial overlap. 
This overlap reflects the randomness of fire occurrence under similar conditions, contributing to noise in classification datasets \citep{kondylatos_probabilistic_2025}.

These various sources of uncertainty can undermine the reliability and interpretability of DL models, thereby limiting their trustworthiness. 
Consequently, it is crucial to quantify both epistemic and aleatoric uncertainties to address all these factors. 
This study responds to this need by presenting a case study on short-term wildfire danger forecasting using an uncertainty-aware DL framework that jointly models both uncertainty types.
Furthermore, challenges inherent to wildfire danger forecasting---such as data sparsity, distributional shifts, noisy labels, and the demand for reliable predictions---are common across other natural hazard domains. 
As such, the proposed framework holds promise for broader applicability in uncertainty-aware modeling across other natural hazard domains.

\section{Methods}
\label{sec:method}

\subsection{Epistemic Uncertainty Estimation}
\label{sec:epistemic}

We capture epistemic uncertainty using BNNs.
BNNs are a class of stochastic NNs that place probability distributions over model weights.
Given a dataset $(x,y)$, a prior distribution $p(w)$ is placed over the weights $w$, and the posterior distribution $p(w|x,y)$ is computed using the Bayes' theorem:
$$p(w|x,y) = \frac{p(y|x,w)p(w)}{\int p(y|x,w)p(w) dw} \propto p(y|x,w)p(w),$$
where $p(y|x,w)$ denotes the likelihood.

The predictive distribution for a new input pair $(x^*,y^*)$  is obtained by marginalizing over the posterior:
\begin{equation}
\begin{aligned}
p(y^*|x^*) = \int p(y^*|x^*, w)p(w|x,y) dw,
\label{integral}
\end{aligned}
\end{equation}
capturing epistemic uncertainty that arises from variances in model parameter distributions.

Computing the posterior is typically intractable, necessitating the use of approximate inference methods such as MCMC and VI for its calculation.
Yet, given that an approximate posterior is available, the predictive distribution in Eq. \ref{integral} can be approximated using MC sampling.
Specifically, a set of $N$ weight samples $\{w^i\}_{i=1}^N$ is drawn from $p(w|x,y)$.
The predicted output is estimated as $$p = \frac{1}{N} \sum_{i=1}^N p_i,$$
where $p_i = p(y^*|x^*, w^i)$ is the prediction for sample $w^i$.
The epistemic uncertainty can be quantified using the variance of the predicted outputs:
\begin{equation}
\begin{aligned}
\sigma^2 & = \frac{1}{N} \sum_{i=1}^N \left ( p_i - p \right)^2, 
\end{aligned}
\end{equation}
In this study, to approximate the integral in Eq. \ref{integral}, we use three methods: i) Bayes by Backpropagation (BBB) \citep{blundell_weight_2015}, MC Dropout \citep{gal_dropout_2016}, and DEs \citep{lakshminarayanan_simple_2017}.
Both BBB and MC Dropout rely on VI to learn an approximation of the posterior during their training and perform Bayesian model averaging by sampling from the variational distribution.
BBB leverages the reparameterization trick to enable efficient gradient-based optimization using standard backpropagation.
MC Dropout interprets dropout \citep{srivastava_dropout} as approximate Bayesian inference and estimates uncertainty by performing multiple stochastic forward passes during inference.
DEs offer an alternative, non-Bayesian approach by training multiple NNs with the same architecture but with different random initializations. 
The ensemble of predictions is then used to perform Bayesian model averaging.
Although they do not explicitly model a posterior distribution, recent work \citep{wilson_case_2020} has shown that DEs can be interpreted as drawing from a multimodal posterior, with each ensemble member representing a different mode. 
Further details for all methods are provided in Supplementary Material (SM)-\ref{sec:sup1}.

\subsection{Aleatoric Uncertainty Estimation}
\label{sec:aleatoric}

To estimate aleatoric uncertainty, we use the approach proposed by \citet{collier_simple_2020}, which accounts for the heteroscedastic label noise.
This method is well-suited for wildfire danger forecasting, where the annotation process exhibits high stochasticity and considerable label noise \citep{prapas_deep_2021, kondylatos_probabilistic_2025}.
The method places a Gaussian distribution over the logits $u_{c}(x)$ of a softmax classifier such that $u_{c} (x) \sim \mathcal{N} (f_{c}^{w}(x), \sigma_{c}^{w}(x)^2)$, with $c = 1, \ldots, K$, where both $f_c^w(x)$ and $\sigma_c^w(x)$ are predicted for each class $c$.
The predictive distribution $p_c$ is approximated using a temperature-scaled softmax:
\begin{equation}
\begin{aligned}
    p_c & \approx \mathbb{E}_{\epsilon_c \sim \mathcal{N}(0, \sigma_c^w(x)^2)}\left[\frac{\exp{(u_{c}(x)/\tau)}}{\sum_{k=1}^K\exp {(u_{k}(x)/\tau)}}\right], \tau > 0.
\end{aligned}
\label{eq:tempered-softmax}
\end{equation}
To approximate this expectation, MC sampling is used,
drawing $S$ samples $\mu_c\sim\mathcal{N}(0,1)$, and compute $u_c^s(x) = f_c^w(x) + \sigma _c^w(x)\mu_c^s$.
The softmax probability for each sample is given by: $$p_{c}^s=\frac{\exp(u_c^s(x)/\tau)}{\sum_{k=1}^{K}\exp(u_k^s(x)/\tau)}.$$
For a new sample $(x^*, y^*)$, the predicted output is estimated as $$p_c(y^*|x^*) = p_c = \frac{1}{S}\sum_{s=1}^{S} p_{c}^s.$$
The corresponding aleatoric uncertainty is calculated as: $$\sigma_c^2 = \frac{1}{S} \sum_{s=1}^S (p_{c}^s - p_c)^2.$$
More details on the method can be found on SM-\ref{sec:sup2}.

\subsection{Total Uncertainty Estimation}

In the preceding sections, we described methods for calculating aleatoric and epistemic uncertainty independently.
In this section, we present a unified framework that enables the joint modeling of both uncertainties.
This approach leverages one of the BNN architectures presented in Sec.~\ref{sec:epistemic} to estimate epistemic uncertainty, and the method presented in Section \ref{sec:aleatoric} to estimate aleatoric uncertainty.
The full inference process is described in Alg.~\ref{algorithm}.

We integrate the aleatoric uncertainty module into the Bayesian formulation of Eq. \ref{integral}.
For this, we replace the standard softmax with the temperature-scaled approximation of Eq. \ref{eq:tempered-softmax}.
Consequently, the predictive distribution marginalizes not only over the posterior of the model weights but also over the noise $\epsilon_c$, which accounts for input-dependent aleatoric uncertainty in the logit space.

The final predictive distribution for class $c$ is calculated as:  
\begin{equation}
\begin{aligned}
    p_c(y^*|x^*) & = \int \mathbb{E}_{\epsilon_c \sim \mathcal{N}(0, \sigma_c^w(x)^2)}p(y^*|x^*,w) p(w|x,y) \ dw\\
    & = \int \int p(y^*, \epsilon_c|x^*,w) \ d\epsilon_c \ p(w|x,y) \ dw \\
    & = \int \int p(y^*|x^*,w, \epsilon_c) p(\epsilon_c|x^*,w) \ d\epsilon_c \ p(w|x,y) \ dw \\ 
\end{aligned}
\end{equation} 
To approximate this nested integral, a \textit{double MC estimation} is employed:
\begin{itemize}
\item We sample a set of $N$ weights $\{w^i\}_{i=1}^N$ from the approximate posterior $p(w|x,y)$.
\item For each $w^i$, we draw $S$ samples of the logit noise $\epsilon_c \sim \mathcal{N}(0, \sigma_c^{w^i}(x)^2)$  (practically from $\mu_c \sim N(0,1)$) to estimate the final output probabilities $p_c^{i, s}$.
\end{itemize} 

The predictive probability for each class $c$ is approximated by averaging over both stochastic weight samples and logit noise samples: 
$$p_c = \frac{1}{N} \sum_{i=1}^N \frac{1}{S}\sum_{s=1}^{S} p_c^{i, s},$$ where $$p_c^{i, s} = \frac{\exp((f_c^{w^i}(x) + \sigma_{c}^{w^i}(x)\mu_{c}^{s})/\tau)}{\sum_{k=1}^{K}\exp((f_{k}^{w^i}(x) + \sigma_{k}^{w^i}(x)\mu_{k}^{s})/\tau)}$$
The total uncertainty ($TU_c$) for class $c$ is estimated as the variance of all $N\times S$ MC samples: $$ TU_c = \frac{1}{N} \sum_{i=1}^N \frac{1}{S}\sum_{s=1}^{S} (p_c^{i,s} - p_c)^2.$$

\begin{figure}
    \centering
    \includegraphics[width=0.7\linewidth]{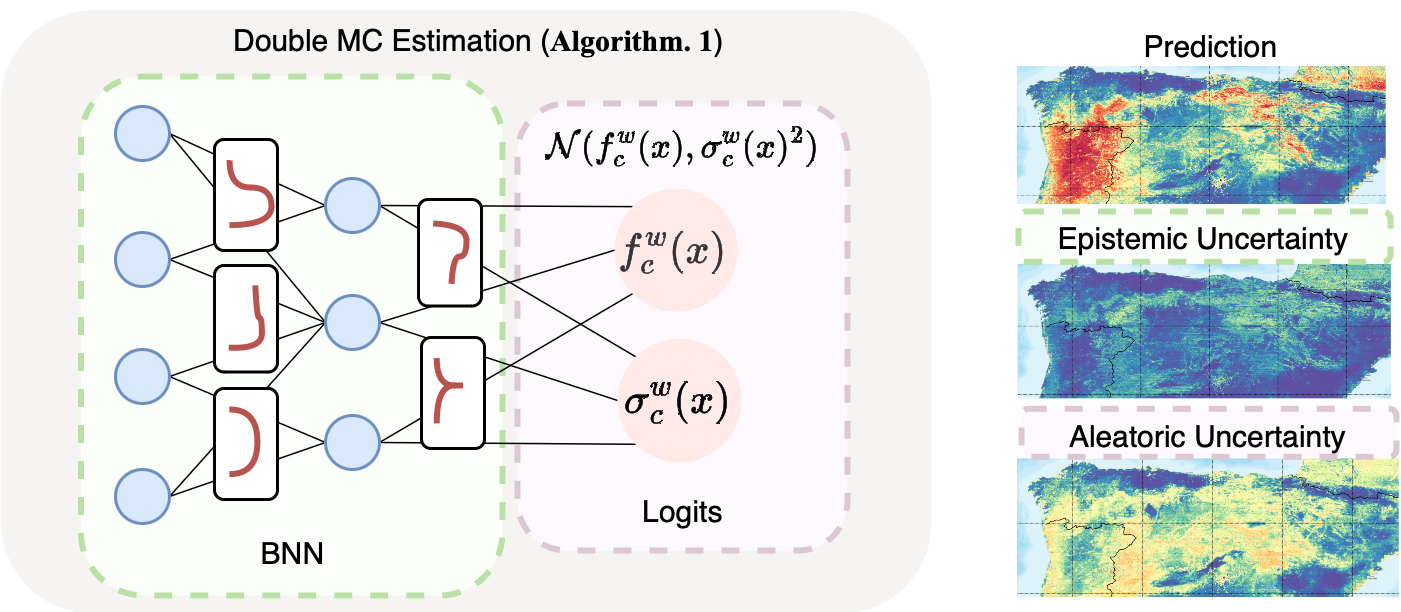}
    \caption{Overview of the proposed framework for estimating total uncertainty. The epistemic uncertainty is modeled using a Bayesian Neural Network, which is combined with a probabilistic module to capture aleatoric uncertainty. The final predictive distribution is obtained through double Monte Carlo (MC) sampling, as described in Alg.~\ref{algorithm}.}
    \label{fig:method}
\end{figure}

\begin{algorithm}[H]
	\caption{Estimation of total uncertainty}
    \label{algorithm}
    \begin{small}
	\begin{algorithmic}
        \State Let $N$ be the MC samples for approximating the posterior of the weights
        \State Let $S$ be the MC samples for the approximation of the normal distribution of the probabilistic framework
        \State Let $EU_c$ be epistemic uncertainty, $AU_c$ be aleatoric uncertainty, and $TU_c$ be total uncertainty.
		\For {$i = 1, \ldots N$}
        \For {$s = 1, \ldots S$}
        \State Sample $\mu_c^s \sim N(0,1)$
		\State Calculate $p_c^{i, s} = \frac{\exp((f_c^{w^i}(x) + \sigma_{c}^{w^i}(x)\mu_{c}^{s})/\tau)}{\sum_{k=1}^{K}\exp((f_{k}^{w^i}(x) + \sigma_{k}^{w^i}(x)\mu_{k}^{s})/\tau)}$
        \State Add $p_c^{i, s}$ in a list
        \EndFor
        \State Calculate mean $\bar{p_c^{i}} = \frac{1}{S}\sum_{s=1}^{S} p_c^{i, s}$
        \State Calculate variance $\sigma_c^{2^{i}} = \frac{1}{S}\sum_{s=1}^{S} (p_c^{i, s} - \bar{p_c^{i}})^2$
        \State Add $\bar{p_c^{i}}$ in a list
        \State Add $\sigma_c^{2^{i}}$ in a list
		\EndFor
        \State Calculate prediction $p_c = \frac{1}{N} \sum_{i=1}^N \frac{1}{S}\sum_{s=1}^{S} p_c^{i, s}$,
		\State Calculate $EU_c =\frac{1}{N} \sum_{i=1}^N (\bar{p_c^{i}} - p_c)^2$, 
        \State Calculate $AU_c = \frac{1}{N} \sum_{i=1}^N \frac{1}{S} \sum_{s=1}^S \sigma_c^{2^{i}}$
        \State Calculate $TU_c = EU_c + AU_c$
        \State Empty all lists
	\end{algorithmic} 
    \end{small}
\end{algorithm}

To disentangle the total uncertainty into its epistemic and aleatoric components, we isolate the contributions arising from uncertainty in model parameters and logit noise, respectively.

\noindent\textit{Epistemic uncertainty for class $c$ ($EU_c$)}: For each weight sample $w_i$, we compute the mean prediction over its $S$ logit noise samples: $$\bar{p_c^{i}} = \frac{1}{S}\sum_{s=1}^{S} p_c^{i, s}.$$
Epistemic uncertainty is then estimated as the variance of these $N$ mean predictions:
$$EU_c = \frac{1}{N} \sum_{i=1}^N (\bar{p_c^{i}} - \frac{1}{N}\sum_{i=1}^N\bar{p_c^{i}})^2 =\frac{1}{N} \sum_{i=1}^N (\bar{p_c^{i}} - p_c)^2.$$ 
This formulation captures the uncertainty attributable solely to the variability in model weights, marginalizing over the logit noise.

\noindent \textit{Aleatoric uncertainty for class $c$ ($AU_c$)}: For each weight sample $w^i$, we compute the variance over its $S$ logit noise samples.
We then average these variances across all $N$ weight samples:
\begin{equation}
\begin{aligned}
AU_c & = \frac{1}{N}\sum_{i=1}^N \frac{1}{S}\sum_{s=1}^S (p_c^{i,s} - \frac{1}{S}\sum_{s=1}^{S} p_c^{i, s}) \\
 & = \frac{1}{N} \sum_{i=1}^N  \underbrace{\frac{1}{S} \sum_{s=1}^S (p_c^{i, s} - \bar{p_c^{i}})^2.}_{\text{aleatoric for sample $w_i$}}
\end{aligned}
\end{equation}
This formulation isolates the uncertainty introduced by logit-level noise, independently of model parameter uncertainty, and averages it over all sampled models.
Apart from the intuitive explanation presented in the main text, we provide an analytical proof that this decomposition satisfies $TU_c = EU_c + AU_c$ (SM-~\ref{sec:sup3}).

\section{Experiments \& Evaluation Metrics}
\label{sec:experiments}

The experimental analysis is organized into two parts. 
Section \ref{sec:next_day} presents the core analysis on next-day wildfire danger forecasting.
For this task, we train three categories of model variants, each modeling a different type of uncertainty, as outlined in Sec. \ref{sec:method}: (i) a model that captures aleatoric uncertainty using the probabilistic framework; (ii) models that estimate epistemic uncertainty using BBB, DEs, and MC Dropout; and (iii) models that estimate both aleatoric and epistemic uncertainty by integrating the aleatoric uncertainty module into the epistemic uncertainty-aware models.
These models are compared against a baseline deterministic DL model that does not provide uncertainty estimates.
The analysis includes: (i) assessing predictive performance and calibration; (ii) evaluating the reliability of uncertainty estimates; and (iii) examining the disentanglement of epistemic and aleatoric uncertainty.
Section~\ref{sec:different_temporal} provides a complementary analysis of wildfire danger prediction across multiple forecasting horizons, examining how model performance and uncertainty evolve as the prediction horizon increases. 

For each of the 10-day forecast horizons, we use a temporally distinct DL model specifically tasked to predict wildfire danger for the fixed target day $t$ using inputs from $n$ days in advance, where $n \in {1, 2, \ldots, 10}$.
For each lead time $n$, a distinct dataset is created by extracting a 45-day window from the original 55-day time series, spanning days $ t-n-44$ to $ t-n$.
This procedure yields 10 unique datasets, corresponding to the respective forecast horizons, each of which is used to train a dedicated model. 
Notably, the target labels remain consistent across all models and lead times, indicating whether a wildfire event occurred or not at the target pixel on the fixed prediction day $t$.
The case of $n = 1$ corresponds to next-day fire danger forecasting.
All experiments use a Long Short-Term Memory (LSTM) architecture, with training procedures adapted to support uncertainty estimation where applicable.
Models are trained using a cross-entropy loss function, weighted by the size of the burned area to emphasize the impact of larger fire events.
Details on model architecture and training hyperparameters are provided in SM-\ref{sec:sup_total}.

Model performance is evaluated using standard classification metrics: Precision, Recall, F1 score, and the Area Under the Precision-Recall Curve (AUPRC). 
To assess calibration, i.e., the agreement between predicted probabilities and the actual likelihood of outcomes---we use reliability diagrams and compute the Expected Calibration Error (ECE) \citep{guo_calibration}.  
The reliability of uncertainties is evaluated through three complementary approaches:
(1) The Discard Test \citep{haynes_creating_2023}, which progressively removes the most uncertain predictions and monitors how average model loss changes in each iteration; a reliable model should demonstrate a decrease in error as more uncertain samples are discarded.
(2) Uncertainty Density Plots \citep{stahl_evaluation_2020}, which visualize the distribution of uncertainties for correctly and incorrectly classified samples; for a model to be reliable, misclassified samples should exhibit higher uncertainty than correctly classified ones.
(3) The AUROC and AUPRC between predicted uncertainty scores and prediction correctness---a binary value indicating whether the model’s prediction is correct or not.
This approach is inspired by the R-AUROC metric that evaluates the reliability of uncertainties in the representation space \citep{kirchhof2023url}. 
Reliable models are expected to achieve high AUROC and AUPRC values, indicating a strong negative correlation between uncertainty and predictive correctness (e.g., low correctness should correspond to high uncertainty and vice versa).
Further details on these metrics are provided in SM-\ref{sec:sup4}.

\section{Results}

\subsection{Next-day wildfire danger forecasting}
\label{sec:next_day}

This section focuses on predicting wildfire danger one day in advance.
We train both uncertainty-aware DL models and a baseline deterministic DL model, as detailed in Section~\ref{sec:experiments}.
The datasets and the code are available in \url{https://github.com/Orion-AI-Lab/uncertainty-wildfires}

\subsubsection{Evaluation of model performance}

\begin{table*}[htbp]
    \caption{Performance comparison across uncertainty-aware models with and without aleatoric uncertainty (AU) compared to the deterministic baseline. Arrows indicate whether higher ($\uparrow$) or lower ($\downarrow$) values are preferred for each metric. ECE refers to Expected Calibration Error. The best results for each metric are highlighted in bold.}
    \label{tab:results}
    \centering
    \resizebox{\textwidth}{!}{%
    \setlength{\tabcolsep}{12pt}
    \begin{tabular}{ccccccc}
        \toprule
        Model & AU & Precision ($\uparrow$) & Recall ($\uparrow$) & $F1$ ($\uparrow$) & AUPRC ($\uparrow$) & ECE ($\downarrow$) \\
        \cmidrule(lr){1-7}
        Deterministic & No & 0,784 & 0,741 & 0,762 & 0,840 & 0,031 \\
        \cmidrule(lr){2-7}
        Aleatoric-only & Yes & 0,789 & 0,751 & 0,770 & 0,846 & 0,022 \\
        \cmidrule(lr){2-7}
        \multirow{2}{*}{Monte Carlo Dropout} & No & 0,783 & 0,763 & 0,773 & 0,840 & 0,023 \\
                                    & Yes & 0,787 & 0,755 & 0,771 & 0,842 & 0,016 \\
                                    \cmidrule(lr){2-7}
        \multirow{2}{*}{Deep Ensembles} & No & \textbf{0,794} & 0,760 & 0,777 & 0,852 & 0,019 \\
                             & Yes & 0,787 & 0,768 & 0,777 & \textbf{0,854} & 0,022 \\
                             \cmidrule(lr){2-7}
        \multirow{2}{*}{Bayes By Backpropagation} & No &  0,776 & 0,782 & 0,779 & 0,852 & 0,012 \\
                             & Yes & 0,753 & \textbf{0,820} & \textbf{0,785} & \textbf{0,854} & \textbf{0,010} \\
        \bottomrule
    \end{tabular}%
    }
\end{table*}

Table \ref{tab:results} summarizes the performance of all models.
Models incorporating uncertainty---whether capturing aleatoric, epistemic, or both---consistently outperform the deterministic baseline in terms of both F1 Score and AUPRC. 
The most significant performance gains are observed when adding the epistemic uncertainty component, highlighting the effectiveness of BNNs in modeling the various sources of epistemic variability inherent to this task (Sec.~\ref{sec:uncertainty_need}). 
Furthermore, the integration of aleatoric uncertainty leads to additional, albeit smaller, improvements across all BNN variants. 
In line with the findings of \citet{kondylatos_probabilistic_2025}, we hypothesize that this performance gain arises from the model's ability to capture the intrinsic noise in the data. 

The most notable performance gain is observed with the BBB model trained with the aleatoric uncertainty module, which achieves an F1 Score 2.3\% and an AUPRC 1.4\% higher than that of the deterministic DL baseline. 
Notably, this model improves Recall by 7.9\%, indicating a substantially enhanced ability to correctly identify the actual wildfire events. 
This improvement, however, comes at the expense of reduced Precision, reflecting an increase in false positives. 
Such a trade-off is often acceptable in high-stakes applications, such as wildfire forecasting, where the cost of missing true events (false negatives) is higher than that of generating false alarms \citep{begueria_validation_2006}.
Moreover, this concern is alleviated by the model's ability to maintain a good balance between Precision and Recall, as reflected in its high F1 score, while also avoiding overconfidence, as indicated by its low ECE.

\subsubsection{Evaluation of model calibration}

\begin{figure}
  \centering
    \includegraphics[width=0.7\linewidth]{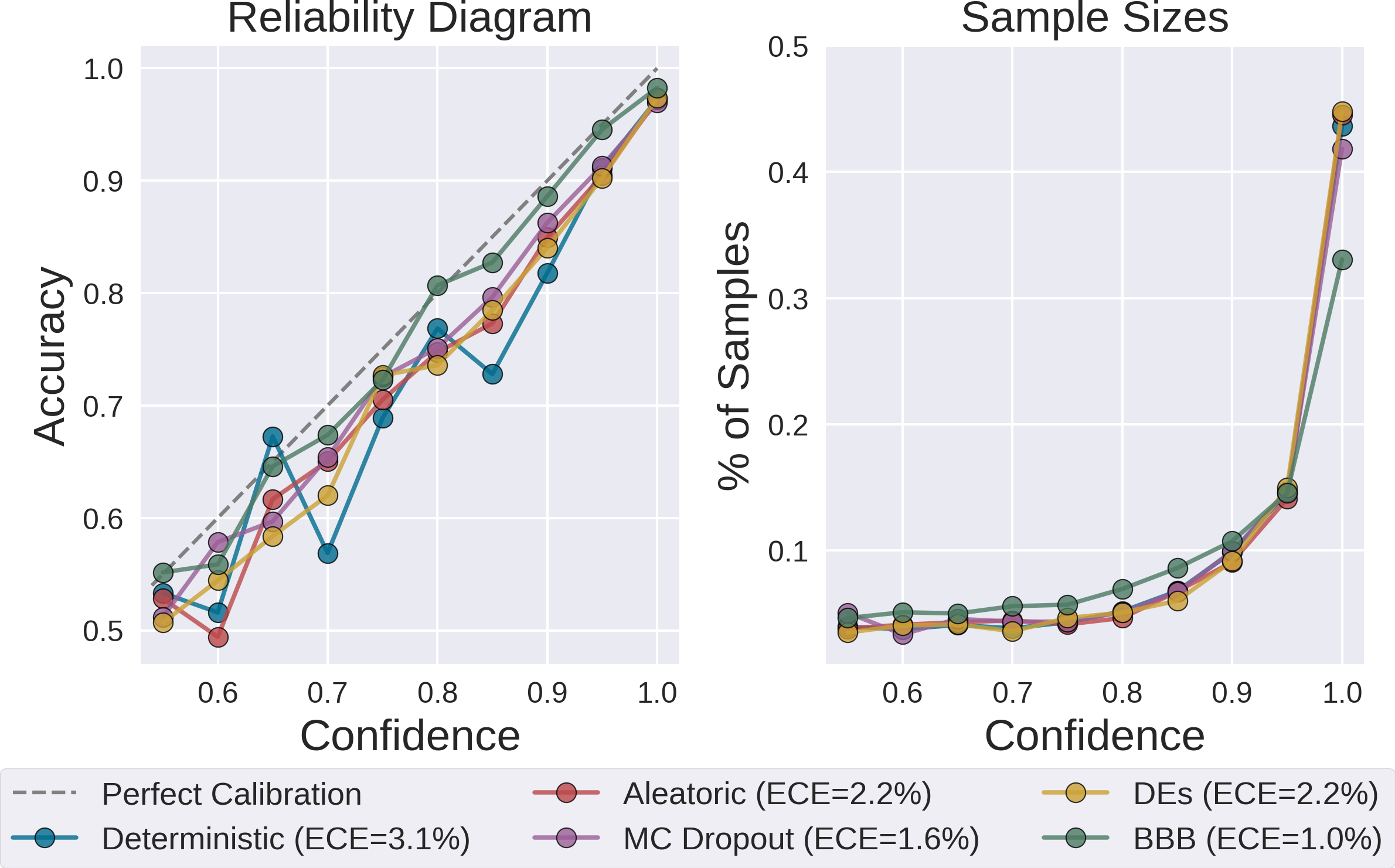}
    \caption{Left: Reliability diagrams for all models trained with aleatoric uncertainty. A well-calibrated model should align closely with the diagonal, indicating strong agreement between predicted confidence and observed accuracy. Right: Distribution of samples across confidence bins in the reliability diagrams.}
    \label{fig:calibration}
\end{figure}

Table \ref{tab:results} also reports the ECE for all models.
All methods that incorporate uncertainty demonstrate significantly better calibration than the deterministic model.
In particular, the BBB model trained with the aleatoric uncertainty module achieves the best calibration, yielding an ECE value, which is lower by 2.1 \% compared to the deterministic baseline.
It is important to note that the improvement in calibration is achieved without the use of any post-hoc calibration techniques such as temperature scaling or Platt's scaling \citep{guo_calibration}, but solely through the inherent properties of uncertainty-aware DL modeling.

To further analyze model calibration, we present reliability diagrams for all methods in Fig. \ref{fig:calibration}.
Each diagram is accompanied by a line plot showing the percentage of test samples falling into each confidence bin. 
The main text focuses on BNN variants trained with the aleatoric uncertainty module, while diagrams for models without aleatoric uncertainty are provided in Fig.~\ref{fig:sup_calibration}.
These visualizations offer insight into whether models are overconfident (confidence scores below the diagonal) or underconfident (confidence scores above the diagonal). 
A perfectly calibrated model would align with the diagonal, indicating that predicted confidence levels match exactly the observed accuracy for each bin.
Since wildfire danger prediction is formulated as a binary classification task, all maximum predicted class probabilities are above $0.5$, resulting in an absence of samples in the lower-confidence bins.
Thus, the diagrams are shown only for $ \geq 0.5$ bins.

The results from the reliability diagrams support the findings from the ECE metrics, revealing that the deterministic model exhibits the poorest calibration, being consistently overconfident across all bins. 
Furthermore, as demonstrated in the sample diagram, a substantial percentage of its predictions are concentrated in the highest confidence bins (near 1.0). 
In contrast, the BBB model---both with and without aleatoric uncertainty---exhibits the best calibration behavior. 
Its reliability curve closely follows the diagonal, indicating a strong alignment between predicted confidence and observed accuracies across the entire confidence spectrum.
This yields well-calibrated predictions for both low- and high-confidence cases.
Moreover, its confidence distribution is more evenly spread across bins, avoiding the overaccumulation of predictions near 1.0, which reflects a more conservative and better-calibrated predictive behavior. 
The remaining models exhibit intermediate calibration performance, with DEs and aleatoric-only models exhibiting worse behavior than MC Dropout in terms of both curve alignment and confidence distribution.

A crucial question that arises is how predictive skill varies with confidence level, and, more specifically, which model exhibits the highest predictive skill when it is most confident.
Figure~\ref{fig:sup_conf_levels} presents the F1 Score and AUPRC of each model across confidence bins from 0.5 to 1.0 (in increments of 0.1).
The BBB model, both with and without aleatoric uncertainty, achieves the highest performance across all confidence levels.
Moreover, the performance gap widens as confidence levels increase.
Notably, in the highest confidence bin ($>0.9$), where the model expresses the highest confidence, the BBB model surpasses the deterministic baseline by up to 5\% in the F1 Score.
This indicates that the BBB model not only produces better-calibrated probabilities but also translates confidence into actual predictive skill.

\subsubsection{Evaluation of model uncertainties}

Evaluating uncertainty is an essential part when developing uncertainty-aware models, particularly in operational contexts where unreliable or misinformed predictions can be misleading. 
To assess the reliability of uncertainties, we employ the Discard Test, Uncertainty Density Plots, and calculate alignment between uncertainty and prediction correctness as detailed in Sec. \ref{sec:experiments}.
The main text includes results for BNN models trained with the aleatoric uncertainty module, while results for those trained without it are provided in the SM.
For each model, the reported uncertainty reflects its total estimated uncertainty: aleatoric for aleatoric-only models, epistemic for epistemic-only models, and both for models trained to capture both epistemic and aleatoric uncertainty.

\begin{figure}[htbp]
  \centering
    \includegraphics[width=0.5\linewidth]{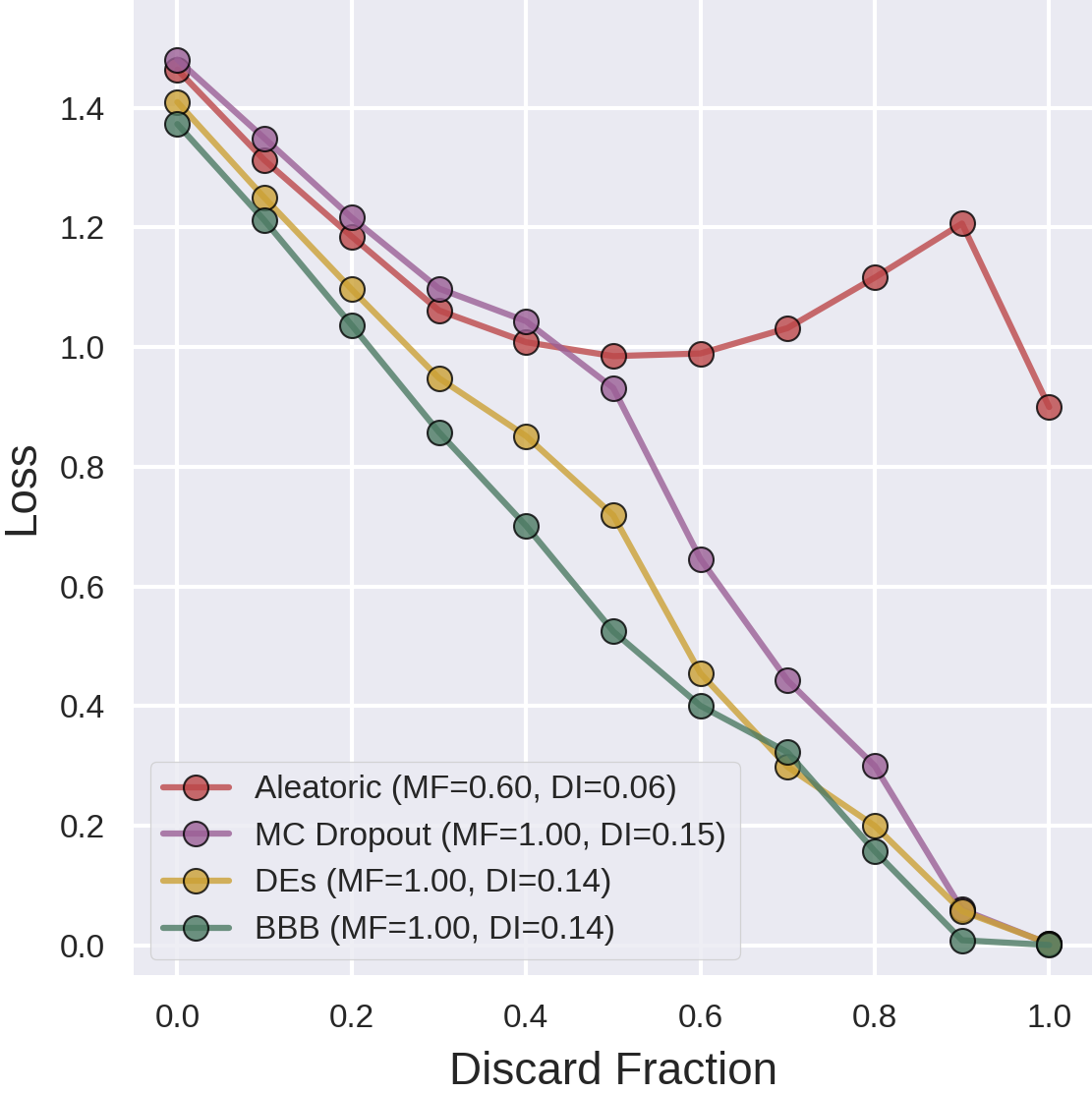}
    \caption{Discard test plots for all models trained with aleatoric uncertainty using the loss as error measurement. A reliable model should exhibit a decreasing loss as the discard fraction increases, indicating that the most uncertain samples correspond to higher loss values. MF (Monotonicity Fraction) indicates how often the loss decreases when samples are removed, while the Discard Improvement (DI) quantifies the average loss reduction as the discard fraction increases.}
    \label{fig:discard}
\end{figure}

\paragraph{Discard Test} 
The Discard Test progressively removes the least certain samples and monitors the change in average loss to evaluate whether the uncertainties align with model errors. 
A trustworthy model should show a decrease in loss as the discard fraction increases, suggesting that the most uncertain samples are associated with higher loss values. 
For models that account for epistemic uncertainty (Fig.~\ref{fig:discard}), we consistently observe a decline in loss as the discard fraction increases, indicating a clear relationship between high uncertainty and high-loss samples.
For example, discarding the top 20\% most uncertain samples leads to a reduction in average loss of 0.33 for BBB, 0.26 for MC Dropout, and 0.31 for DEs for the models trained with aleatoric uncertainty.
A similar trend is observed for models without aleatoric uncertainty (Fig.~\ref{fig:sup_discard_loss}), though with slightly lower improvements: 0.29 for BBB, 0.23 for MC Dropout, and 0.34 for DEs.
These findings have important implications for operational decision-making, as discussed in Section~\ref {sec:decision_making}.
The model trained to account only for aleatoric uncertainty exhibits an initial decrease in loss, but a slight increase in the later discard fractions indicates less reliable uncertainty estimation.
This highlights the contribution of epistemic uncertainty to more robust estimates of total uncertainty.
Moreover, all BNN variants demonstrate an MF of 1.0, confirming that loss consistently decreases with increasing discard fraction. The DI of 0.15 and 0.14 reflect a significant average loss reduction in each step. 

In Fig.~\ref{fig:sup_discard_f1} and ~\ref{fig:sup_discard_auprc}, we replicate the discard analysis using F1 Score and AUPRC as performance metrics, replacing the loss used in the main analysis. 
A reliable uncertainty estimate is expected to yield higher performance metrics when the least confident predictions are discarded. 
Results indicate that all models improve initially, with the BBB model showing the most pronounced gains. 
For instance, discarding the top 20\% most uncertain samples improves the BBB model’s F1 Score by 7.5\%, and this increase reaches 15\% when 60\% of the most uncertain samples are removed, leading to an overall score of 93.6\%. 
All BNN variants exhibit a steep drop in performance after a certain amount of discarded fractions, unlike the aleatoric-only model, which remains more stable.
This decline can be attributed to the low number of positive samples in the retained subsets, as shown in the inset plots depicting the percentage of positive instances across different discard fractions. Given the strong class imbalance, even a small number of errors in the minority class can substantially impact the metrics.
The consistent drop in the percentage of positives as the discard fraction increases also implies that BNNs assign higher uncertainty to positive samples, in contrast to aleatoric-only models, which are more balanced in this aspect.
This indicates that fire events are more uncertain epistemically, likely reflecting a greater variability in their input space compared to non-fire events. This characteristic cannot be captured by the aleatoric-only model, which does not account for epistemic uncertainty.

\begin{figure}
\centering
\includegraphics[width=0.7\linewidth]{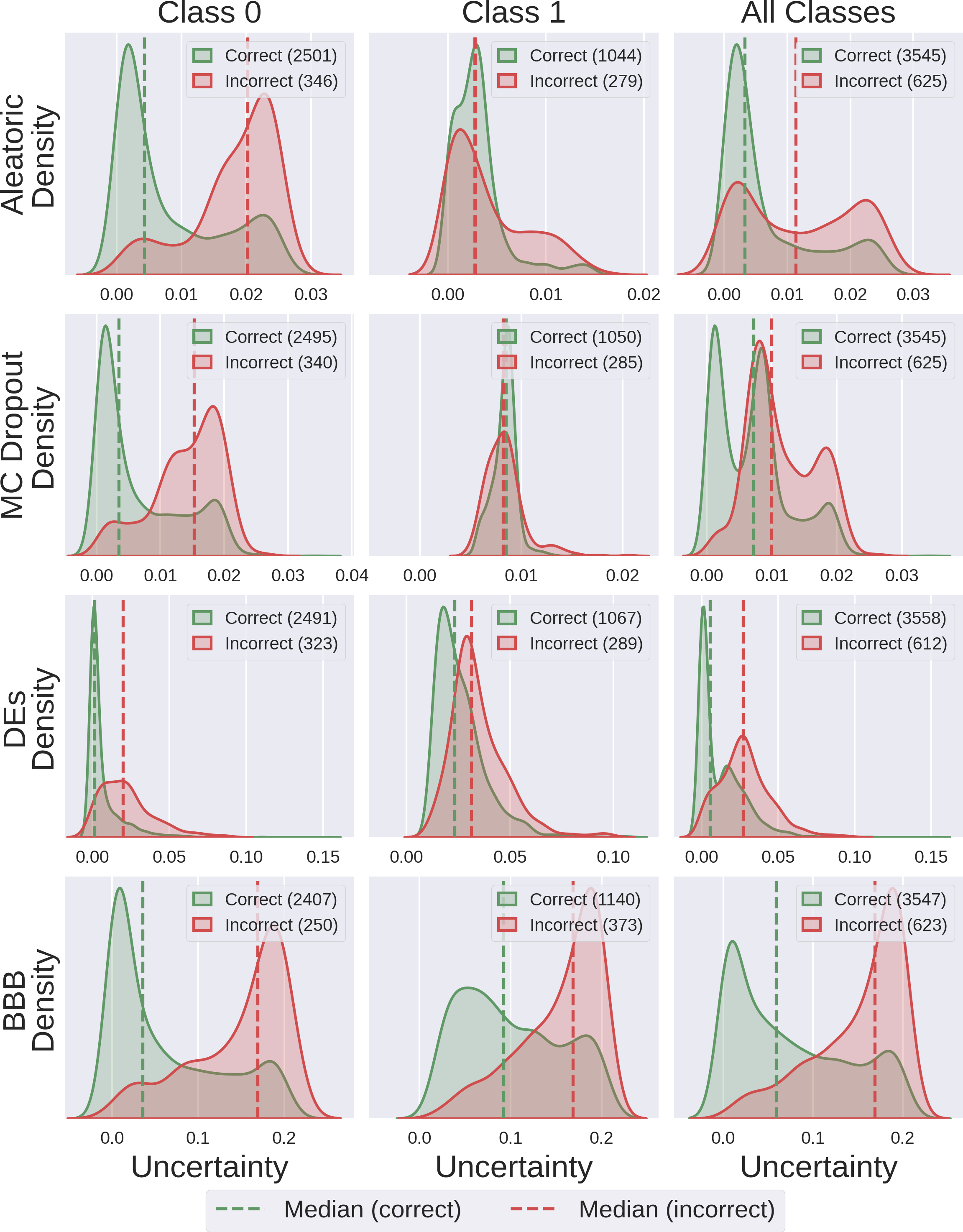}
  \caption{Uncertainty density plots across all models trained with aleatoric uncertainty. The plots are presented for the two classes combined as well as separately for the positive and negative classes. Vertical dashed lines mark the median uncertainty for each group. Clearly separated distributions with minimal overlap characterize reliable uncertainty estimates.}
  \label{fig:densities}
\end{figure}

\paragraph{Uncertainty Density Plots} The uncertainty density plots (Figures~\ref{fig:densities} and ~\ref{fig:sup_density}) offer additional insights into the estimated uncertainties, as they evaluate the models' ability to differentiate the uncertainties between correctly and incorrectly classified samples, both across all classes and for each class separately.
For the DEs and BBB models, the uncertainty density plots demonstrate a clear separation between correctly and incorrectly classified samples. 
Correct predictions consistently show uncertainty distributions that lean towards lower values, while misclassified samples are linked to higher uncertainty. Specifically, the BBB model displays a noticeable peak of high uncertainty for misclassified samples, whereas predictions with low uncertainty tend to be accurate.
This behavior aligns with the reliability criteria proposed by \citet{mukhoti_evaluating_2019}, which suggests that low-uncertainty estimates should correspond to accurate predictions, while inaccurate predictions should correspond to high-uncertainty values.

The model trained only with aleatoric uncertainty, as well as the MC Dropout model, performs well in separating class 0 but fails to distinguish effectively between correctly and incorrectly classified samples in class 1. 
For the aleatoric uncertainty model, misclassified samples exhibit a bimodal distribution when considering all classes, suggesting that the model fails to assign appropriately high uncertainty to its errors almost 50\% of the time.
MC Dropout performs even worse in this regard, as the distributions of correct and incorrect predictions substantially overlap beyond class 0. 
The confusion is particularly evident in the combined classes plot, where several overlapping modes can be seen. These findings suggest two key points: firstly, including epistemic uncertainty is crucial for enhancing the reliability of total uncertainty estimates in this context; and secondly, straightforward methods like MC Dropout are inadequate for accurately capturing significant uncertainty in complex scenarios such as wildfire danger forecasting.
This observation is consistent with prior findings in classical computer vision literature, where MC Dropout has also been shown to produce unreliable uncertainty estimates \citep{gustafsson_evaluating_2020}.
An additional insight is that uncertainty ranges differ across models, implying that thresholds for interpreting high and low uncertainty are model-dependent.

\begin{table}[ht]
\centering
\caption{AUROC and AUPRC scores between predicted uncertainty and prediction correctness, a binary indicator of whether the model’s prediction is correct. Higher values indicate that the model’s uncertainty estimates effectively correlate with prediction errors, reflecting better reliability. The best values for each metric are highlighted in bold. }
\label{tab:uncertainty_auroc_auprc}
\resizebox{0.6\textwidth}{!}{%
\setlength{\tabcolsep}{12pt}
\begin{tabular}{llcccc}
\toprule
Model & AU & AUROC & AUPRC \\
\cmidrule(lr){1-4}
Aleatoric & Yes & 0.658 & 0.889 \\
\cmidrule(lr){2-4}
\multirow{2}{*}{MC Dropout} & No & 0.723 & 0.942\\
                            & Yes & 0.716 & 0.938 \\
\cmidrule(lr){2-4}
\multirow{2}{*}{DEs}        & No & 0.794 & 0.958  \\
                            & Yes & 0.785 & 0.956 \\
\cmidrule(lr){2-4}
\multirow{2}{*}{BBB}        & No & 0.781 & 0.955  \\
                            & Yes & \textbf{0.809} & \textbf{0.960} \\
\bottomrule
\end{tabular}%
}
\end{table}

\paragraph{AUROC and AUPRC between uncertainty and prediction correctness}
Table~\ref{tab:uncertainty_auroc_auprc} evaluates how well the predicted uncertainties align with prediction correctness, using AUROC and AUPRC as metrics.
The BBB model, when combined with aleatoric uncertainty, demonstrates the highest alignment, indicating that its uncertainty estimates are the most accurate predictors of model correctness. 
DEs also perform well, especially in their configuration that considers only epistemic uncertainty; however, their performance slightly declines when aleatoric uncertainty is included. In contrast, both the aleatoric-only model and MC Dropout display weaker alignment.
These findings are consistent with the results from uncertainty density plots, underscoring the importance of modeling epistemic uncertainty and the advantages of using more expressive Bayesian methods for improving the reliability of wildfire danger forecasting.

\subsubsection{Disentanglement of epistemic and aleatoric uncertainty}

Recent work has raised concerns regarding the ability of uncertainty estimation methods to effectively disentangle epistemic and aleatoric uncertainty in computer vision tasks, showing that most existing approaches produce highly correlated estimates for the two uncertainty types \citep{muscanyi_neurips_2024}. 
Other works advocate for the independent quantification of each uncertainty source using distinct modeling techniques \citep{mukhoti_2023_cvpr}.
Building on this reasoning, our approach specifically uses distinct methods to estimate epistemic and aleatoric uncertainty. In this section, we assess whether these estimates are independent in the context of forecasting wildfire danger or if they contain overlapping information.

\begin{figure}
  \centering
    \includegraphics[width=0.9\linewidth]{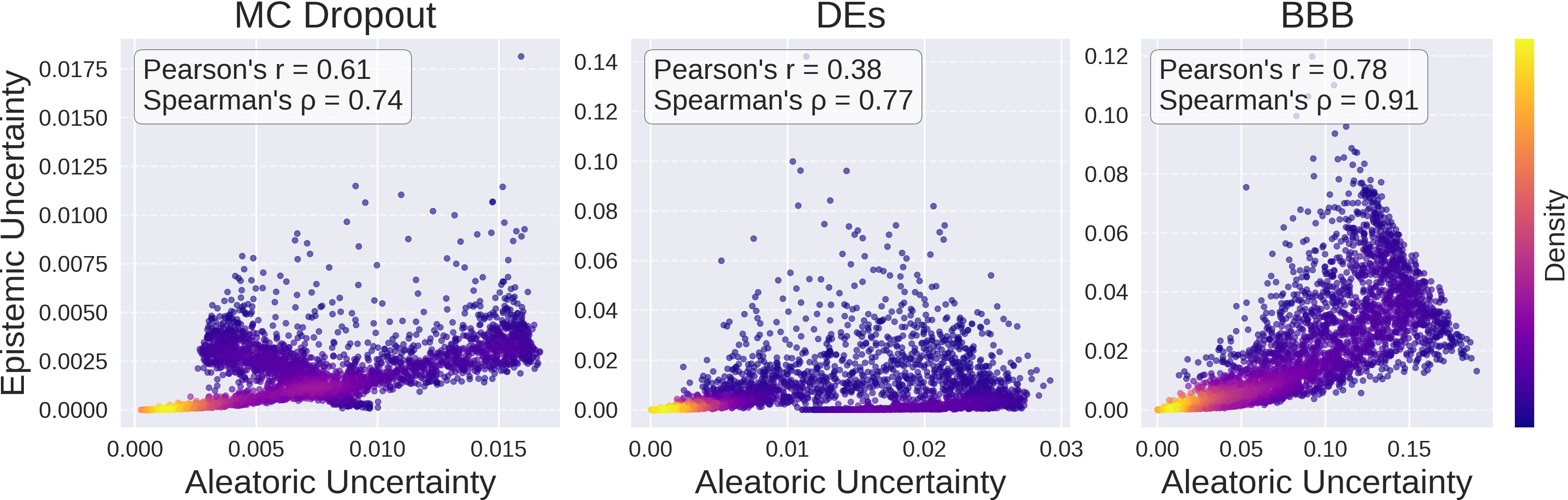}
    \caption{Correlation between aleatoric uncertainty (x-axis) and epistemic uncertainty (y-axis) for all models trained to estimate both types. Pearson and Spearman correlation coefficients are provided for each model. The color bar represents sample density, with higher color intensity indicating regions of greater point overlap.}
    \label{fig:scatterplot}
\end{figure}

Figure~\ref{fig:scatterplot} presents scatter plots that illustrate the relationship between aleatoric and epistemic uncertainty across all models that account for both.
As observed, each model produces varying correlation patterns, reflecting distinct ways of capturing and expressing uncertainty. 
Yet, for all models, most samples exhibit low epistemic uncertainty---suggesting sufficient model knowledge---and a strong concentration in regions characterized by both low aleatoric and epistemic uncertainty, as indicated by the yellow coloration in these areas.
Additionally, Pearson's and Spearman's correlation coefficients between aleatoric and epistemic uncertainty are typically high, particularly for the BBB model, where Spearman's rank correlation reaches 91\%.

These preliminary findings indicate a strong alignment between the learned aleatoric and epistemic uncertainties for most samples; that is, both tend to increase or decrease together. However, the interpretive value of this correlation varies based on the difficulty of the sample. 
In easy-to-classify cases, such as clear negatives during periods of low fire risk—where uncertainty is consistently low—the strong correlation between the two types of uncertainty provides limited additional insights. On the other hand, in more complex or ambiguous scenarios characterized by high total uncertainty, distinguishing between the epistemic and aleatoric components becomes more valuable, as they can offer complementary perspectives on model confidence and data noise.

\begin{table}[htbp]
\centering
\caption{Spearman and Pearson correlation coefficients between aleatoric and epistemic uncertainties calculated when progressively removing samples with the highest total uncertainty, based on percentile thresholds. This analysis highlights how the correlation changes as increasingly uncertain samples are excluded.}
\label{tab:correlations}
\resizebox{0.6\textwidth}{!}{%
\begin{tabular}{llcccc}
\toprule
Model & Percentile & Spearman & & Pearson & \\
\midrule
\multirow{4}{*}{MC Dropout} 
 & Full data   & 0.741 & & 0.609 & \\
 & $> 25^{th}$ & 0.387 & & 0.404 & \\
 & $> 50^{th}$ & 0.238 & & 0.268 & \\
 & $> 75^{th}$ & 0.380 & & 0.113 & \\
\midrule
\multirow{4}{*}{DEs} 
 & Full data   & 0.766 & & 0.379 & \\
 & $> 25^{th}$ & 0.458 & & 0.268 & \\
 & $> 50^{th}$ & -0.047 & & -0.075 & \\
 & $> 75^{th}$ & -0.635 & & -0.488 & \\
\midrule
\multirow{4}{*}{BBB} 
 & Full data   & 0.912 & & 0.777 & \\
 & $> 25^{th}$ & 0.800 & & 0.690 & \\
 & $> 50^{th}$ & 0.464 & & 0.370 & \\
 & $> 75^{th}$ & -0.555 & & -0.591 & \\
\bottomrule
\end{tabular}
}
\end{table}

To understand whether the large number of low-uncertainty samples mainly causes the strong correlation between uncertainties, we perform an additional analysis. 
We gradually remove the most uncertain samples, based on total uncertainty percentiles, and recalculate the correlation between aleatoric and epistemic uncertainty at each step.
Results for this experiment are summarized in Table~\ref{tab:correlations}.
The findings show that correlation decreases substantially as increasingly uncertain samples are removed.
Notably, for DEs and BBB models, correlation drops below $-0.5$ after excluding samples above the $75^{th}$ percentile, indicating a shift toward negative correlation in the most uncertain samples.
This reduction in correlation indicates that the models have learned to distinguish between epistemic and aleatoric uncertainties as distinct sources of information, particularly as the samples become more challenging. 

\subsection{Predicting at different temporal horizons}
\label{sec:different_temporal}

\begin{figure}
    \centering
    \includegraphics[width=0.6\linewidth]{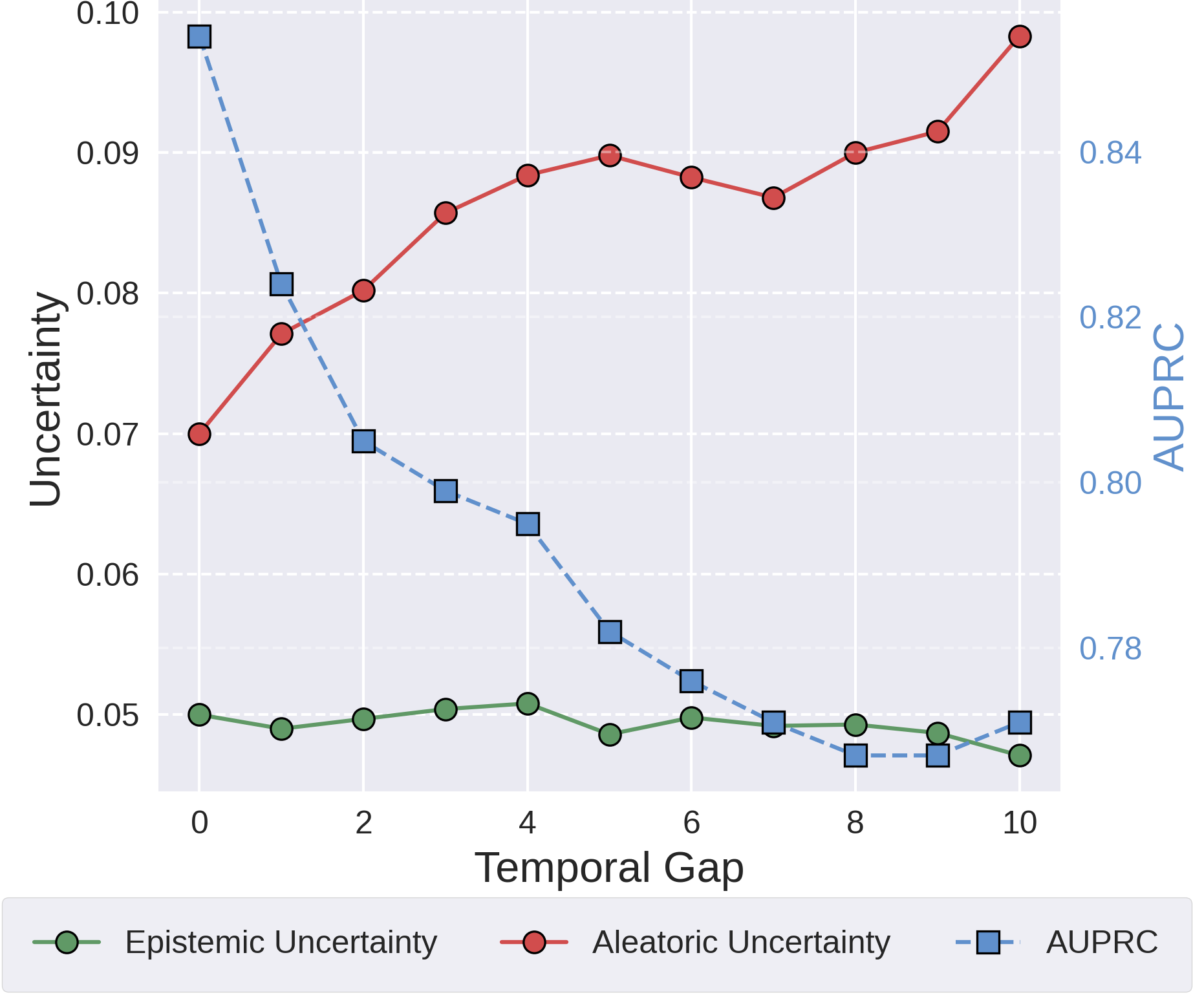}
    \caption{Evolution of model performance, measured by AUPRC, alongside average aleatoric and epistemic uncertainty values for wildfire danger predictions from 1 to 10 days ahead. The x-axis indicates the prediction lead time, while the left y-axis shows uncertainties and the right y-axis shows the AUPRC.}
    \label{fig:temporal_horizons}
\end{figure}

In this section, we extend our analysis to investigate how model performance and uncertainty estimates evolve when forecasting wildfire danger at varying lead times from 1 to 10 days in advance (Figure \ref{fig:temporal_horizons}).
For this analysis, we use the BBB model trained with the aleatoric uncertainty module, as it demonstrated the most reliable performance in earlier evaluations.
The figure presents the prediction lead time on the x-axis, while the two y-axes correspond to the model’s AUPRC, used as the performance metric indicator, and the average aleatoric and epistemic uncertainty values across all test samples.

A first observation is that the model's performance, measured by AUPRC, declines gradually as the forecasting horizon increases. 
This decline aligns with expectations, as predicting wildfire danger becomes more complex when forecasting further into the future. 
However, the degradation in performance is relatively modest as the AUPRC drops by only 8\% between the 1-day and the 10-day forecast models, indicating that the model retains considerable predictive skill even at longer forecasting horizons.
While having DL models that accurately predict wildfire danger several days in advance is valuable, our primary objective is to analyze how aleatoric and epistemic uncertainties behave as the forecasting horizon increases.

First, we observe a systematic increase in aleatoric uncertainty with longer prediction lead times. 
Since aleatoric uncertainty captures the inherent and irreducible noise in the data, this increase suggests that the input features become progressively noisier and less informative for the model when predicting fire danger several days ahead. 
In the context of wildfire forecasting, this is likely due to the increasing stochasticity of fire-driving variables, such as meteorological conditions and vegetation dynamics, in the days preceding fire ignition.
These variables exhibit high variability that cannot be reliably learned from temporally distant observations, thus becoming progressively less representative of the fire danger of the target day.
This naturally reduces the model's ability to confidently associate the values of input features with wildfire danger, leading to increased data uncertainty.

In contrast, epistemic uncertainty remains remarkably stable across all forecasting horizons. 
Epistemic uncertainty reflects the model’s lack of knowledge, often arising from insufficient data or limited model capacity. 
Its stability across lead times suggests that the selected training process---a separate model trained for each forecasting horizon using distinct training datasets---has equipped each model with sufficient information to generalize within its respective temporal horizon.

\section{Discussion}
\label{sec:discussion}

\subsection{Advantages of uncertainty-aware modeling}

Uncertainty-aware DL models consistently outperformed the deterministic DL model in both predictive accuracy and model calibration.
More importantly, beyond improved performance, these models provided an additional layer of insight by estimating uncertainty. 
The uncertainty estimates generated by these models were consistent with predictive errors across various experimental settings, demonstrating their reliability. Additionally, separating uncertainty into epistemic and aleatoric components provided a deeper understanding of the predictions. For example, when making predictions over different time horizons, we found that aleatoric uncertainty increased with lead time, which indicated the rising noise in longer-range wildfire predictions, while model uncertainty remained stable. 

Our results also indicated that the task is characterized by a higher degree of aleatoric uncertainty compared to epistemic uncertainty, as evident from Fig.~\ref{fig:scatterplot} and Fig. \ref{fig:temporal_horizons}. 
This suggests a significant portion of the predictive uncertainty arises from the inherent noise in the task itself and is irreducible within the current supervised classification framework. 
The relatively low levels of epistemic uncertainty across models imply that the proposed approaches are well-specified and generalize effectively across varying conditions.
Yet, although the aleatoric uncertainty is higher, the Discard Test analysis showed that the fire class is more epistemically uncertain than the non-fire class.
This demonstrates the potential benefits of incorporating more fire-labeled samples during training, which could enable the model to better capture the underlying, complex mechanisms that drive fire occurrence.

In all experiments, the BBB models consistently demonstrated superior performance compared to other models.
While purely Bayesian DL models such as BBB are recognized for their robust uncertainty quantification and calibration, prior studies in domains such as computer vision have often reported inferior predictive performance compared to deterministic models or alternative BNN methods \citep{ovadia_can_2019}.
However, our results showed that the BBB-based model achieved the strongest performance across all evaluated methods.
As discussed in Sec. \ref{sec:uncertainty_need}, wildfire danger forecasting is a challenging task characterized by distributional shifts in input variables over time and space, limited data availability, and spatio-temporal heterogeneity.
Our experiments demonstrate that BBB’s capability to learn a more comprehensive posterior distribution over network weights allows it to manage complexity more effectively. While BBB generally incurs higher computational costs for both training and inference, its proven improvements in predictive performance, calibration, and uncertainty reliability make it a valuable tool for high-stakes applications, such as forecasting wildfire danger. Therefore, integrating more principled Bayesian methods could be an intriguing avenue for future research and implementation in these critical tasks.

\subsection{Implications in decision-making}
\label{sec:decision_making}

The findings of our study can have some important implications for real-world decision-making in wildfire danger forecasting. 

\paragraph{Calibration} 
In operational contexts, decision-makers rely on wildfire danger forecasts to guide critical actions such as resource allocation, deployment of firefighting personnel, and implementation of preventive measures. 
Well-calibrated models ensure that predicted probabilities align with the true underlying risk, thereby minimizing the likelihood of misguided decisions.
Overestimating risk may lead to unnecessary resource mobilization, increasing operational costs, and placing undue strain on firefighting personnel. 
Conversely, underestimating risk poses even greater risks, as it can result in insufficient preparedness and delayed response to emerging fire threats, leaving areas vulnerable to fire outbreaks.
Our results demonstrate that uncertainty-aware models, particularly the BBB model, provide enhanced calibration, which can substantially improve the quality of high-impact decisions.

\paragraph{Uncertainty thresholds} Having a model that assigns low uncertainty to correct predictions and relates incorrect predictions with high uncertainty can be valuable for operational decision-making.
A practical strategy for incorporating uncertainty into operational decision-making is by using uncertainty thresholds based on discard analysis and uncertainty distributions. 
For instance, using the discard test, assume that an F1 Score of $85\%$ is required for wildfire danger forecasts to be trusted.
As shown in Fig. \ref{fig:sup_discard_f1}, BBB achieves this criterion when the top 20\% most uncertain predictions are discarded.
This allows decision-makers to set an uncertainty cutoff at the uncertainty value corresponding to this discard fraction and rely only on predictions falling below it for operational decisions.
Similarly, uncertainty density plots (Fig. \ref{fig:densities}) can inform thresholds by identifying the median uncertainty associated with either correct or incorrect classifications (depending on the desired strictness).
Predictions exceeding this cutoff can be flagged as unreliable and excluded from automated pipelines.  
It is essential to note that the selective use of predictions based on uncertainty thresholds requires attention. 
For instance, excluding highly uncertain samples may lead to missed detections in vulnerable areas.
Thus, in real-world applications, combining uncertainty-aware forecasts with real-time environmental drivers or expert systems is necessary. 

\begin{figure}
    \centering
    \includegraphics[width=0.75\linewidth]{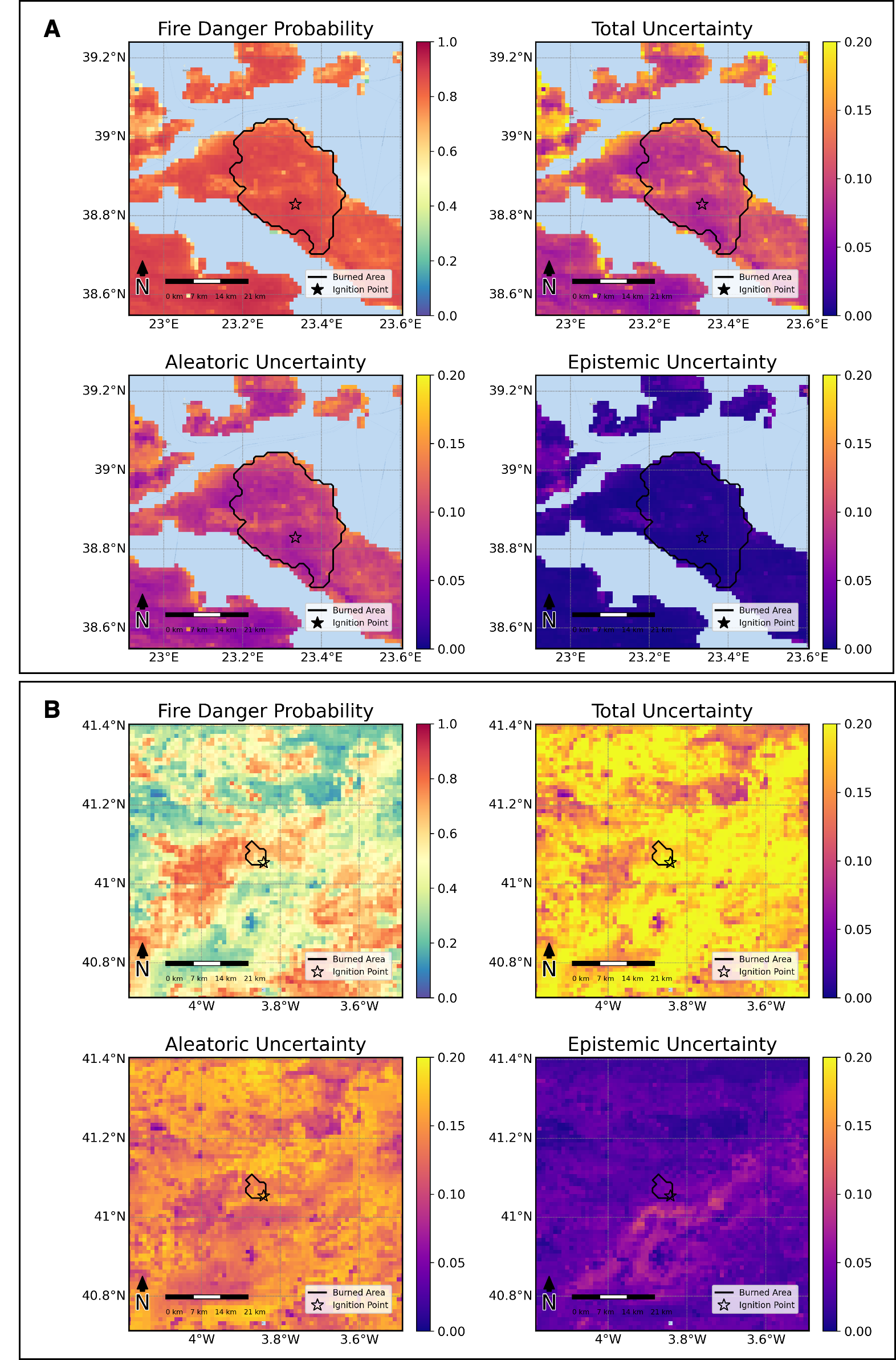}
    \caption{Fire danger maps and associated uncertainties predicted by the BBB model for two wildfire events. (A) Evia, Greece (2 August 2021): The model predicts high fire danger with low uncertainties, indicating high confidence in the prediction. (B) Madrid, Spain (14 July 2022): The predicted danger near the ignition point is relatively low, but the uncertainty high, suggesting the need for increased caution. These maps demonstrate how uncertainty estimates can enhance interpretability and guide decision-making.}
    \label{fig:maps}
\end{figure}

\paragraph{Fire Danger Maps}
Fire danger maps, which include both aleatoric and epistemic uncertainties, enable users to assess not only where fire danger is high, but also the model's confidence in these estimates.
This distinction helps identify inherently unpredictable regions (high aleatoric uncertainty) and those that may benefit from further data or model improvement (high epistemic uncertainty).
The nature of the maps, which can be updated daily and span both local and broader regions, facilitates the identification of persistent patterns or short-term changes in both fire danger and uncertainty, further enhancing interpretability and model understanding. 
Such maps support decision-makers to: i) prioritize areas with high predicted danger and low uncertainty, ii) exercise increased caution in areas with low danger but high uncertainty, and iii) combine DL predictions with human expertise or external data in ambiguous regions.
Figure~\ref{fig:maps} shows two example maps, generated by the BBB model, each tailored to a different decision-making scenario.
In the fire of Evia, Greece (2 August 2021) (Fig.~\ref{fig:maps}-A), where approximately 50,000 hectares were burned, the model predicted high danger with low uncertainty, making it a clear candidate for early intervention. 
The event near Madrid, Spain (14 July 2022) (Fig.~\ref{fig:maps}-B), where 868 hectares were affected, presents a different case: although the predicted danger at the ignition point is relatively low, the uncertainty is high, suggesting increased caution or further investigation.

\subsection{Limitations \& Perspectives}
Despite the promise of our results for more reliable wildfire danger forecasting, we would like to highlight some limitations and underscore existing opportunities for our work.
The modeling approach adopted in this study followed the experimental setup and data provided by Mesogeos, which primarily captured environmental variables while narrowly including factors related to human activities.
Incorporating additional data sources, such as human suppression activities, could help us better understand or even reduce uncertainty. 
Moreover, as discussed in Sec. \ref{sec:uncertainty_need}, the proposed uncertainty-aware framework shows promise for broader application across various natural hazard domains. 
However, each domain introduces its unique sources of uncertainty, which may necessitate domain-specific modeling strategies. 
Therefore, extending the evaluation of uncertainty-aware models to other natural hazards could provide valuable insights into their applicability.
A critical factor influencing the operational utility of such models is how effectively risk is communicated \citep{persello_2021}. 
Historical evidence suggests that even accurate forecasts may fail to prevent disasters if risks are not effectively communicated, understood, and acted upon \citep{camps-valls_artificial_2025}.
While the inclusion of uncertainty estimates can increase trust in AI-driven systems, it also adds complexity, compounding the existing challenges in integrating AI into established decision-making workflows \citep{guikema_artificial_2020}.
Thus, communicating uncertainty to non-technical stakeholders remains a key consideration.

\section{Conclusion}

In this study, we examined the role of uncertainty-aware DL in wildfire danger forecasting, focusing on its potential to enhance predictive performance, trust, and interpretability. 
We developed various DL models that capture both epistemic and aleatoric uncertainty. 
Notably, our model trained with Bayesian Bayes by Backpropagation together with aleatoric uncertainty consistently outperformed the deterministic baseline and other methods like Monte Carlo Dropout across key criteria, including predictive performance and uncertainty reliability.
Our analysis highlighted the complementary roles of epistemic and aleatoric uncertainty, especially in challenging scenarios. While aleatoric uncertainty increased with longer prediction horizons, epistemic uncertainty remained stable.
Finally, we demonstrated several ways in which integrating uncertainty estimates into operational workflows can support more informed decision-making. 
Given that many natural hazard-related tasks share similar challenges, such as data sparsity, distributional shifts, and the need for reliable and interpretable models, the presented framework holds promise for broader applications.
Overall, our work underscores the value of uncertainty-aware DL systems in improving the trustworthiness of predictive models in real-world, high-impact applications.



\section*{Acknowldegments}
\noindent This work has received funding from the European Union's Horizon 2020 Research and Innovation under Grant Agreement no. 101036926 (TREEADS) and the European Union’s Horizon Europe WIDERA Coordination and Support Actions under Grant Agreement no.101159723 (MeDiTwin).




\bibliographystyle{elsarticle-harv} 
\bibliography{bibliography}

\input{supp_mat}

\end{document}

%% file: supp_mat.tex
\renewcommand{\thefigure}{SM\arabic{figure}}
\renewcommand{\thesection}{\Alph{section}}
\setcounter{figure}{0}
\setcounter{section}{0}

\newpage

\begin{center}
\large \textbf{Supplementary Material for \textit{``Uncertainty-Aware Deep Learning for Wildfire Danger Forecasting''}}
\end{center}

\section{Epistemic Uncertainty Estimation}
\label{sec:sup1}

\subsection{Bayes by Backpropagation}

Bayes By Backpropagation (BBB) \citep{blundell_weight_2015} is a Variational Inference technique for training Bayesian Neural Networks (BNNs) in a backpropagation-compatible manner.
It uses a tractable variational posterior distribution $q(w|\theta)$---where $\theta = (\mu, \sigma)$ parameterizes a Gaussian distribution $N(\mu, \sigma^2)$---to approximate the true posterior $p(w|x,y)$.
To find the optimal variational parameters $\theta$, BBB minimizes the Kullback-Leibler (KL) divergence between the variational posterior and the true posterior:
$$KL [q(w|\theta)||p(w|x,y)].$$
This is equivalent to minimizing the negative Evidence Lower Bound (ELBO): 
$$L(\theta) = KL [q(w|\theta)||p(w)] - E_{q(w|\theta)}[\log p(x,y|w)],$$
where $p(w)$ is the prior over weights, and $p(x,y|w)$ is the likelihood.
Since the expectation over the variational posterior in the ELBO is intractable, BBB employs Monte Carlo (MC) sampling to approximate the loss.  
After simple calculations and drawing $N$ MC samples $w^{i} \sim q(w|\theta), \ i=1\ldots N$, the loss function is approximated as:
$$L(\theta) \approx\frac{1}{N}\sum_{i=1}^{n} \log q(w^{i}|\theta) - log p(w^{i}) - \log p(x,y | w^{i}).$$

A key component that enables backpropagation through this stochastic optimization process is the reparameterization trick. 
This reparameterizes the stochastic weights $w \sim q(w|\theta)$ as a deterministic function of a noise variable: $$w = \mu + \sigma \odot \epsilon, \ \epsilon \sim N(0,1).$$
Thus, the stochasticity is introduced via sampling of $\epsilon$, while $\mu$ and $\sigma$ are differential deterministic parameters updated through standard gradient descent.
The simplicity of the involved distributions (e.g., Gaussian priors and variational posteriors) facilitates a closed-form evaluation of the ELBO, making BBB a practical and scalable method for Bayesian DL.

\subsection{Monte Carlo Dropout}

MC Dropout \citep{gal_dropout_2016} is a practical and scalable method for approximate VI in neural networks, offering a variational interpretation of dropout regularization, where the dropout mask is applied at training and \textit{inference} time.
Formally, a variational posterior is defined for each weight matrix $\mathbf{W_i}$ of a layer $i$ as: 
\begin{equation}
\begin{aligned}
\mathbf{W_i} & = \mathbf{M_i} \diag(z_{i}) \\
z_{i,j} & \sim \Bernoulli (p_i),
\end{aligned}
\end{equation}
where $z_i$ represents the dropout mask, $p_i$ the dropout probability of layer $i$, and $\mathbf{M_i}$ denotes the underlying deterministic weight matrix before dropout is applied.
Sampling from the variational distribution $q(w_i)$ is like performing the well-known dropout on layer $i$ in a neural network with weights $(\mathbf{M_i})$. 
MC Dropout approximates the predictive distribution by performing multiple stochastic forward passes through the network at test time, where each pass corresponds to sampling a different dropout mask from the Bernoulli distributions.



\subsection{Deep Ensembles}

Deep Ensembles (DEs) offer a simple yet effective approach for uncertainty estimation in DL models. 
The method entails training multiple NNs independently, each initialized with different random weights. 
The diversity induced by these random initializations leads to the approximation of both predictive performance and model uncertainty.
Let $\{\theta_m\}_{m=1}^{M}$ denote the parameters of $M$ independently trained NNs.
Each model produces a prediction $p_{\theta_{m}}(y|x,\theta_m)$ which can be interpreted as a sample from the Bayesian approximation, as defined in Eq. 1.
The final predictive distribution is obtained by averaging these individual outputs, while the corresponding uncertainty can be estimated through the variance across the ensemble predictions.


\section{Aleatoric uncertainty estimation}
\label{sec:sup2}

The method for aleatoric uncertainty estimation considers a latent variable generative process for the labels \citep{collier_simple_2020}.
The generative process is handled by a latent variable $u_{c}(x)$, which is associated with each class $c$ and input $x$.
This variable is the sum of a deterministic vector $f^w_{c}(x)$ and an unobserved stochastic component $\epsilon_{c}$, expressed as: $u_{c}(x) = f^w_{c}(x) + \epsilon_{c}$.
A label is generated by sampling from $u_{c}(x)$ and taking the $\arg\max$ of all classes, i.e. class $c^*$ is the generated label if $u_{c}(x) \leq u_{c*}(x), \forall c \in 1,\ldots K$, where $K$ is the total number of classes. 

The probability $p_c (x)$ that an input $x$ belongs to class $c$, can be then expressed as
\begin{equation}
\begin{aligned}
  p_c(x) & = P(\arg\max_k u_{k}(x)=c) \\ 
         & = \int \mathbf{1} \left\{ \arg\max_k u_k(x) = c \right\} p(\epsilon_c) \, d\epsilon_c
\end{aligned}
\label{eq:generative}
\end{equation}

Under the presence of input-dependent label noise, the assumption of identically distributed $\epsilon_{c}$ becomes restrictive, as the noise source varies from sample to sample and across different classes.
In such cases, it is necessary to account for differing levels of stochasticity for each sample (heteroscedasticity).
To handle this heteroscedasticity, the method introduces a dependency between the noise terms $\epsilon_{c}$, the input, and classes, breaking the identically distributed assumption. 
Specifically, $\epsilon_{c} \sim \mathcal{N}(0, \sigma^w_{c}(x)^2)$, where $\mathcal{N}$ denotes a Normal distribution, and $\sigma^w_{c}(x)^2$ models noise levels that vary based on the input and class.
Under these assumptions, computing $p_c(x)$ in Eq. ~\ref{eq:generative} is intractable.
However, it can be approximated using a temperature-scaled softmax and MC sampling.
Moreover, to enable gradient-based optimization, $u_{c}(x)$ can be reparameterized as $u_c(x) = f_c^w(x) + \sigma _c^w(x)\mu_c$, where $f_c^w(x)$ and $\sigma _c^w(x)$ are deterministic components and $\mu_c\sim\mathcal{N}(0,1)$.

The final calculation of $p_c(x)$ is obtained as:
\begin{equation}
\begin{aligned}
    p_c(x) & = P(\arg\max_k u_{k}(x)=c) \\
           & \approx \mathbb{E}_{\epsilon_{k}\sim\mathcal{N}(0,\sigma^{w}_{c}(x)^2)}\left[\frac{\exp{(u_{c}(x)/\tau)}}{\sum_{k=1}^K\exp {(u_{k}(x)/\tau)}}\right], \tau > 0 \\
           & \approx \frac{1}{S}\sum_{s=1}^{S}\frac{\exp((f_c^{w}(x) + \sigma_{c}^{w}(x)\mu_{c}^{s})/\tau)}{\sum_{k=1}^{K}\exp((f_{k}^{w}(x) + \sigma_{k}^{w}(x)\mu_{k}^{s})/\tau)},
\end{aligned}
\end{equation} 
where S is the number of MC samples.

\section{Total Uncertainty Equals the Sum of Epistemic and Aleatoric Uncertainty}
\label{sec:sup3}

In this section, we provide a proof that the total predictive uncertainty $TU_c$ can be decomposed into the sum of epistemic uncertainty $EU_c$ and aleatoric uncertainty $AU_c$ as defined in the main text.
Total uncertainty is defined as:
$$ TU_c = \frac{1}{N} \sum_{i=1}^N \frac{1}{S}\sum_{s=1}^{S} (p_c^{i,s} - p_c)^2,$$ 
where the overall predictive mean is given by:
$$p_c = \frac{1}{N}\sum_{i=1}^{N}\bar{p_c^{i}}, \text{ where } \bar{p_c^{i}} = \frac{1}{S}\sum_{s=1}^{S} p_c^{i, s}.$$
The epistemic uncertainty is defined as $$EU_c =\frac{1}{N} \sum_{i=1}^N (\bar{p_c^{i}} - p_c)^2,$$ and the aleatoric uncertainty is defined as $$AU_c = \frac{1}{N}\sum_{i=1}^N \frac{1}{S}\sum_{s=1}^S (p_c^{i,s} - \bar{p_c^{i}}).$$


To show the decomposition $TU_c = EU_c + AU_C$, we apply the identity $\mathrm{Var}(X) = \mathbb{E}[X^2] - (\mathbb{E}[X])^2$ to each variance.

First, since $p_c$ is the mean of all $p_c^{i,s}$ across the full set of $N\times S$ samples, the total uncertainty can be rewritten as: $$TU_c = \frac{1}{NS} \sum_{i=1}^N \sum_{s=1}^{S} (p_c^{i,s} - p_c)^2 = \frac{1}{NS} \sum_{i=1}^N \sum_{s=1}^{S} {p_c^{i,s}}^2 - p_c^2$$

Next, using that $p_c$ is also the mean of $\bar{p_c}^i$ across the $N$ samples, the epistemic uncertainty becomes: 
$$EU_c = \frac{1}{N} \sum_{i=1}^N (\bar{p_c^{i}} - p_c)^2 = \frac{1}{N} \sum_{i=1}^N \bar{p_c^{i}}^2 - {p_c}^2.$$

Similarly, since $\bar{p_c^{i}}$ is the mean of $p_c^{i,s}$ across the $S$ samples, the aleatoric uncertainty for sample $w_i$ is:
$$\frac{1}{S}\sum_{s=1}^S (p_c^{i,s} - \bar{p_c^{i}}) = \frac{1}{S}\sum_{s=1}^S{p_c^{i,s}}^2 - {\bar{p_c^{i}}}^2.$$
Thus, the total aleatoric uncertainty can be computed as:
\begin{equation}
\begin{aligned}
   AU_c & =  \frac{1}{N} \sum_{i=1}^N (\frac{1}{S}\sum_{s=1}^S{p_c^{i,s}}^2 - {\bar{p_c^{i}}}^2)\\
   & = \frac{1}{NS}\sum_{i=1}^N\sum_{s=1}^S{p_c^{i,s}}^2 - \frac{1}{N}\sum_{i=1}^N{\bar{p_c^{i}}}^2
\end{aligned}
\end{equation}

By summing the expressions for $EU_c$ and $AU_c$, we obtain:

$$EU_c + AU_C = \frac{1}{N} \sum_{i=1}^N \bar{p_c^{i}}^2 - {p_c}^2 + \frac{1}{NS}\sum_{i=1}^N\sum_{s=1}^S{p_c^{i,s}}^2 - \frac{1}{N}\sum_{i=1}^N{\bar{p_c^{i}}}^2.$$

Canceling out the term $\frac{1}{N} \sum_{i=1}^N \bar{p_c^{i}}^2$, we have that:

$$EU_c + AU_C = \frac{1}{NS}\sum_{i=1}^N\sum_{s=1}^S{p_c^{i,s}}^2 - {p_c}^2 = TU_c.$$

\section{Model architecture and hyperparameters}
\label{sec:sup_total}

The model consists of a normalization layer, followed by an LSTM layer with $128$ neurons, two fully connected layers with $128$ and $64$ neurons, respectively, and a final softmax output layer. 
All intermediate linear layers are followed by a ReLU activation and are regularized using Dropout with a probability of $0.5$
The integration of Dropout at the intermediate layers is critical for enabling the application of MC Dropout.
We use 50 MC samples for both BBB and MC Dropout, and 1,000 MC samples of logit noise to estimate aleatoric uncertainty during both training and inference. 
For DEs, we train 10 independently initialized models.
The hyperparameter $\tau$ is tuned exclusively on the deterministic model using the validation set and then reused unchanged across all other models.
The best values was found to be $0.2$.

\section{Evaluation Metrics}
\label{sec:sup4}

\paragraph{Calibration}
Calibration refers to the alignment between predicted probabilities and the actual likelihood of outcomes. 
Formally, a model is perfectly calibrated if $\mathbb{P}(\hat{y} = y | \hat{p} = p) = p, \forall p\in[0,1]$,
where $\hat{y}$ is the predicted class, $y$ is the actual class and $\hat{p}$ is the predicted probability.
To assess calibration, we use reliability diagrams, which plot the expected accuracy as a function of model confidence.
Model predictions are grouped into $M$ probability bins of equal size $1/M$ from $0$ to $1$.
For each bin $B_m$, $m \in \{1,2,\ldots M\}$, we compute:
\begin{itemize}
\item Accuracy: $acc(B_m) = \frac{1}{|B_m|} \sum_{i \in B_m} \mathbf{1}(\hat{y_i} = y_i).$
\item Confidence: $conf(B_m) = \frac{1}{|B_m|} \sum_{i \in B_m} \hat{p_i},$
\end{itemize}
Perfect calibration means $acc(B_m) = conf(B_m)$ for all $m \in \{1,2,\ldots M\}$, resulting in a diagonal line on the plot, while and deviation from the diagonal suggests that the model is either overconfident or underconfident. 
Calibration can be  summarized in a single metric, ECE, computed as $$ECE = \sum_{m=1}^M\frac{|B_m|}{n} |acc(B_m) - conf(B_m)|,$$ where $n$ is the total number of samples.
ECE captures the average difference between confidence and accuracy across all bins.

\paragraph{Discard Test}

The Discard Test is a diagnostic tool used to assess the quality of a model’s uncertainty estimates by iteratively removing the most uncertain predictions and measuring the resulting change in model error. 
The fundamental principle behind this test is that if a model’s uncertainty estimates are reliable, the most uncertain predictions should correspond to higher errors; thus, removing them should lead to an improvement in overall model performance.
The exact steps of the test are the following: 
\begin{enumerate}
    \item Model predictions are ranked in descending order based on their associated uncertainty estimates.
    \item The ranked samples are divided into equal-sized batches according to a predefined discard fraction.
    \item The most uncertain batch is removed from the set.
    \item The model’s error is recalculated on the remaining test samples.
    \item Steps 3–4 are repeated iteratively until all samples have been discarded.
\end{enumerate}

This process generates a curve that visualizes how the model error changes as more uncertain predictions are excluded. 
An effective uncertainty estimation method should result in a monotonically decreasing error curve, indicating that the most uncertain samples also tend to have higher errors. 
Deviations from these trends, such as non-monotonic error curves, suggest that the uncertainty estimates are not fully reliable, as removing uncertain predictions does not consistently enhance model performance.
In this study, we use $10$ discard fractions, so the steps are repeated $10$ times for each model.
Moreover, we use the model loss, F1 Score, and AUPRC as measures of error.

Two metrics are also reported: Monotonicity Fraction (MF) and Discard Improvement (DI). 
MF captures the consistency of improvement as more samples are discarded and is defined as:
$$MF = \frac{1}{N_{f}-1}\sum_{i=1}^{N_{f}-1}I(\epsilon_{i}\geq\epsilon_{i+1}),$$
where $\epsilon_i$ represents the model error at $i$-th discard fraction and $N_f$ denotes the number of discarded fractions.
An MF value of $1$ indicates perfect monotonicity.
An ideal uncertainty estimation method would yield a high MF (indicating consistent performance improvement). 
The DI measures the average reduction in error across discard steps:
$$DI=\frac{1}{N_{f}-1}\sum_{i=1}^{N_{f}-1}I(\epsilon_{i} - \epsilon_{i+1}).$$

\paragraph{Uncertainty Density Plots}

Uncertainty Density Plots visualize the distribution of uncertainty scores across test samples, separately for correctly and incorrectly classified instances. 
For each group, the median uncertainty is also reported. 
In a reliable model, misclassified samples should generally exhibit higher uncertainty than correctly classified ones, resulting in a clear separation between the two distributions. 
This separation serves as an indicator of the model’s ability to assign meaningful uncertainty estimates.



\begin{figure}
  \centering
    \includegraphics[width=0.7\linewidth]{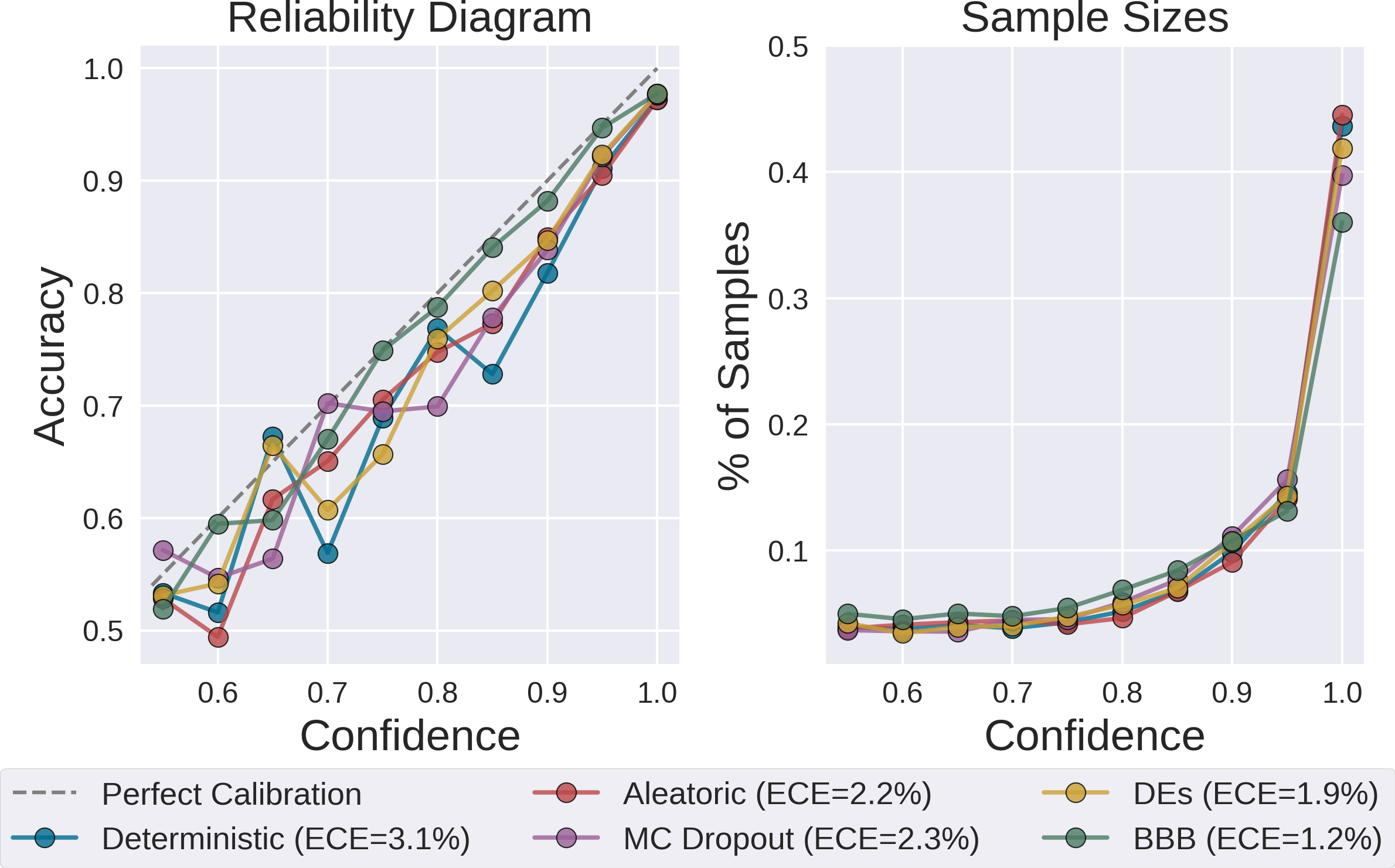}
    \caption{Left: Reliability diagrams for all models trained without aleatoric uncertainty. A well-calibrated model should align closely with the diagonal, indicating strong agreement between predicted confidence and observed accuracy. Right: Distribution of samples across confidence bins in the reliability diagrams.}
    \label{fig:sup_calibration}
\end{figure}

\begin{figure}
  \centering
    \includegraphics[width=0.7\linewidth]{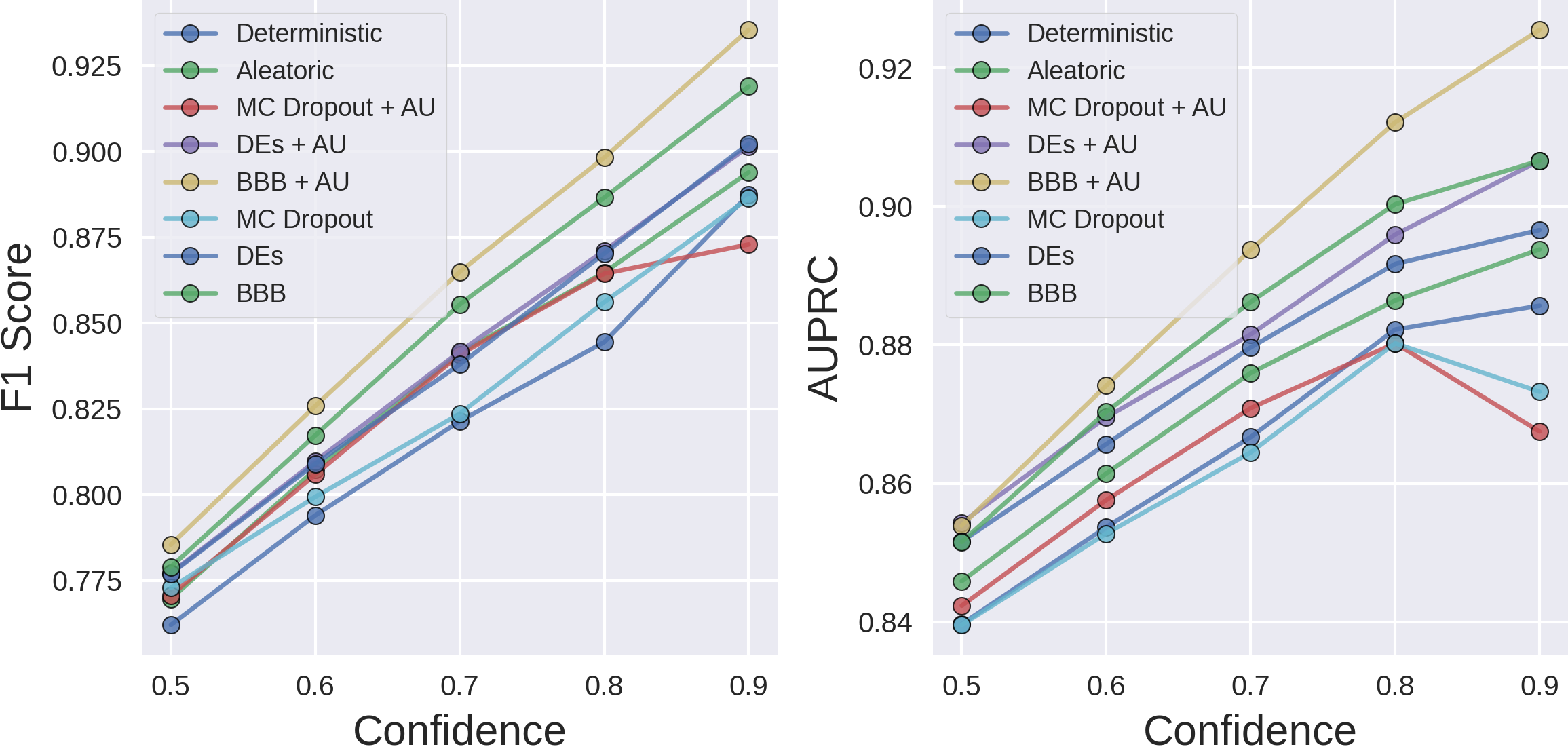}
    \caption{F1 Score and AUPRC of all models across confidence bins.}
    \label{fig:sup_conf_levels}
\end{figure}





\begin{figure}
  \centering
    \includegraphics[width=0.5\linewidth]{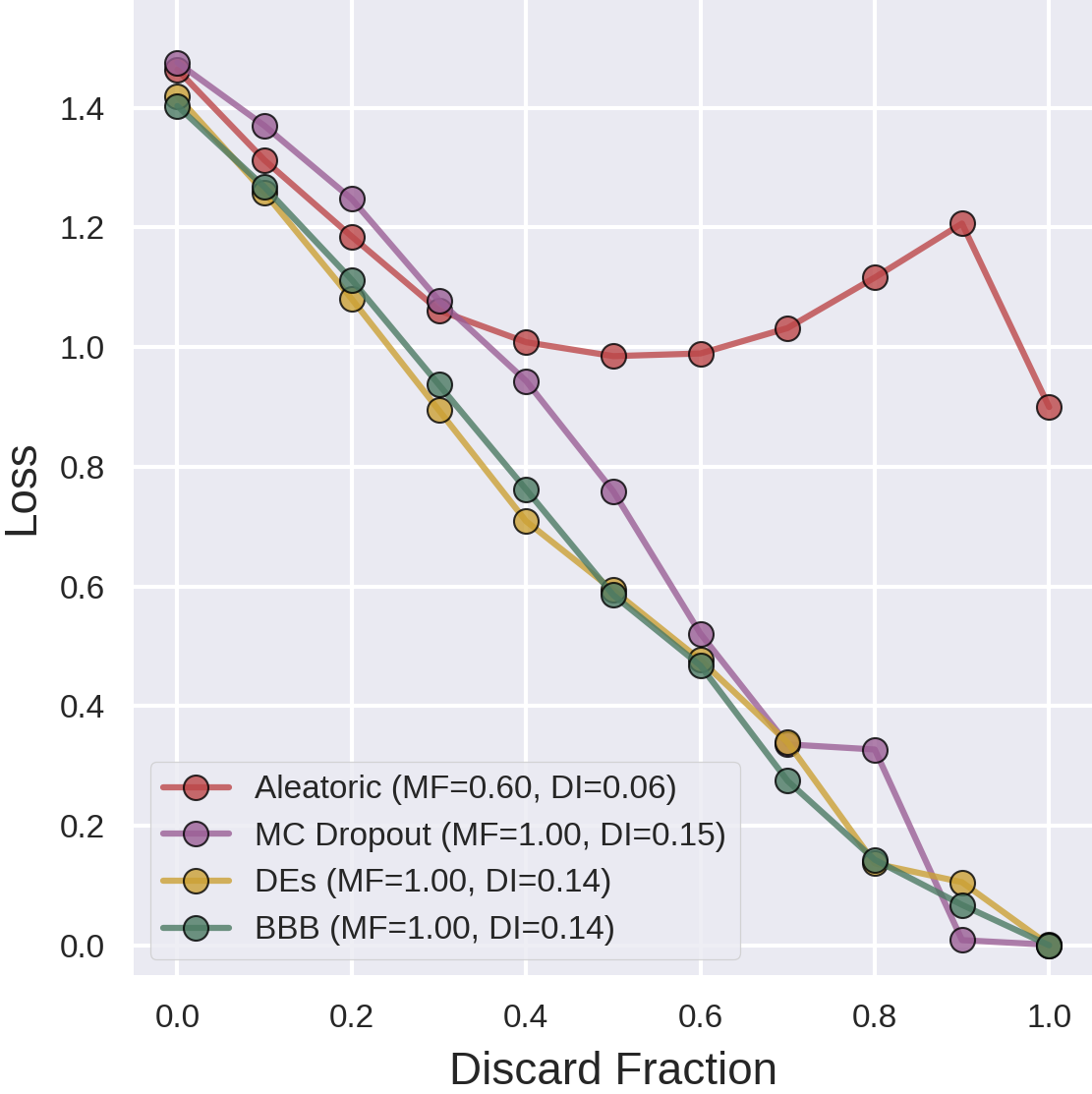}
    \caption{Discard test plots for all models trained without aleatoric uncertainty using the loss as error measurement. A reliable model should exhibit a decreasing loss as the discard fraction increases, indicating that the most uncertain samples correspond to higher loss values. MF (Monotonicity Fraction) indicates how often the loss decreases when samples are removed, while the Discard Improvement (DI) quantifies the average loss reduction as the discard fraction increases.}
    \label{fig:sup_discard_loss}
\end{figure}

\begin{figure}
  \centering
    \includegraphics[width=0.5\linewidth]{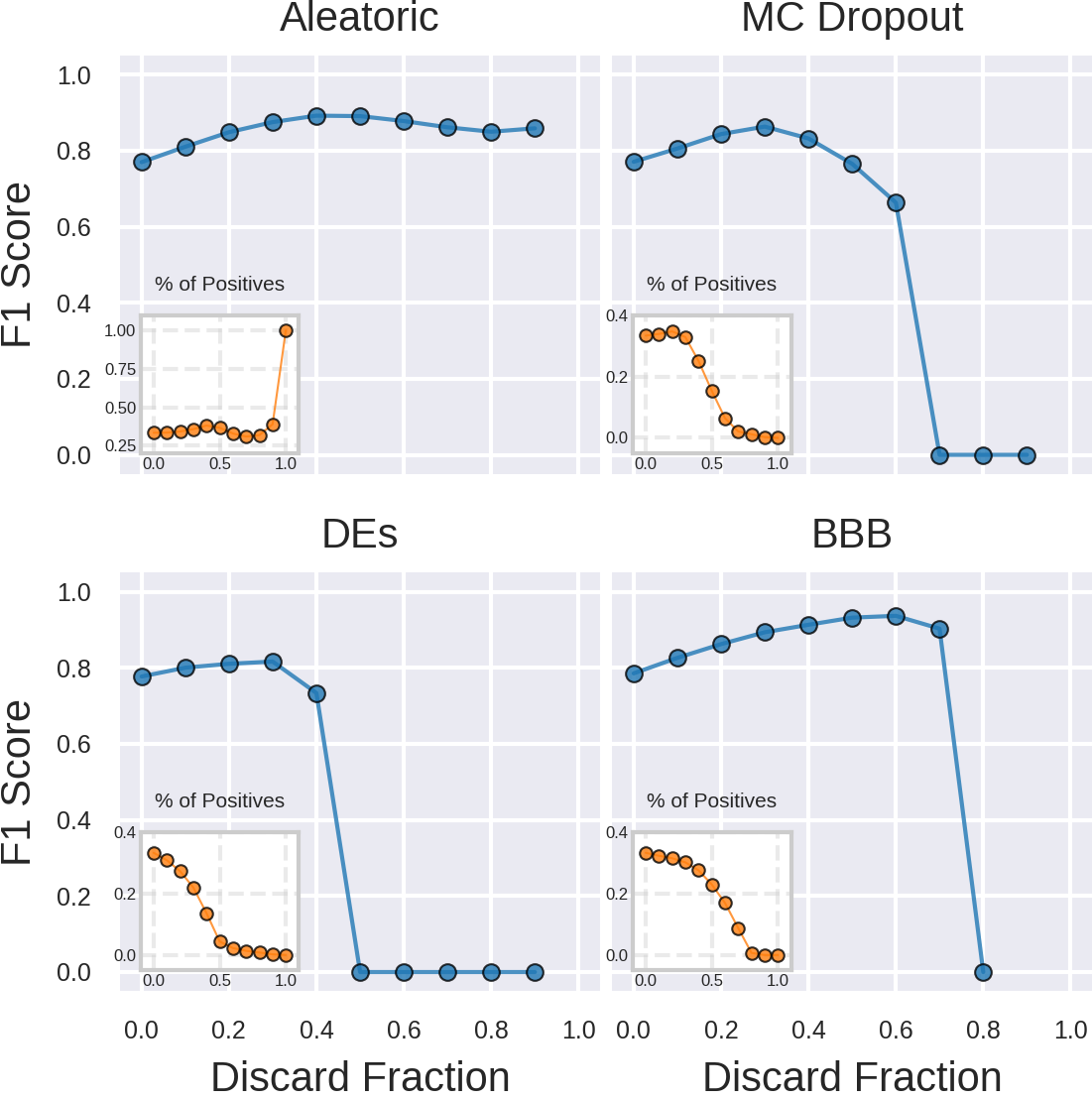}
    \caption{Discard test plots for all models trained with aleatoric uncertainty using the F1 Score as error measurement. A reliable model should exhibit increasing F1 score values as the discard fraction increases, indicating that keeping samples with higher uncertainty corresponds to lower F1 scores. The inset plots display the percentage of positive samples within each discard fraction.}
    \label{fig:sup_discard_f1}
\end{figure}

\begin{figure}
  \centering
    \includegraphics[width=0.5\linewidth]{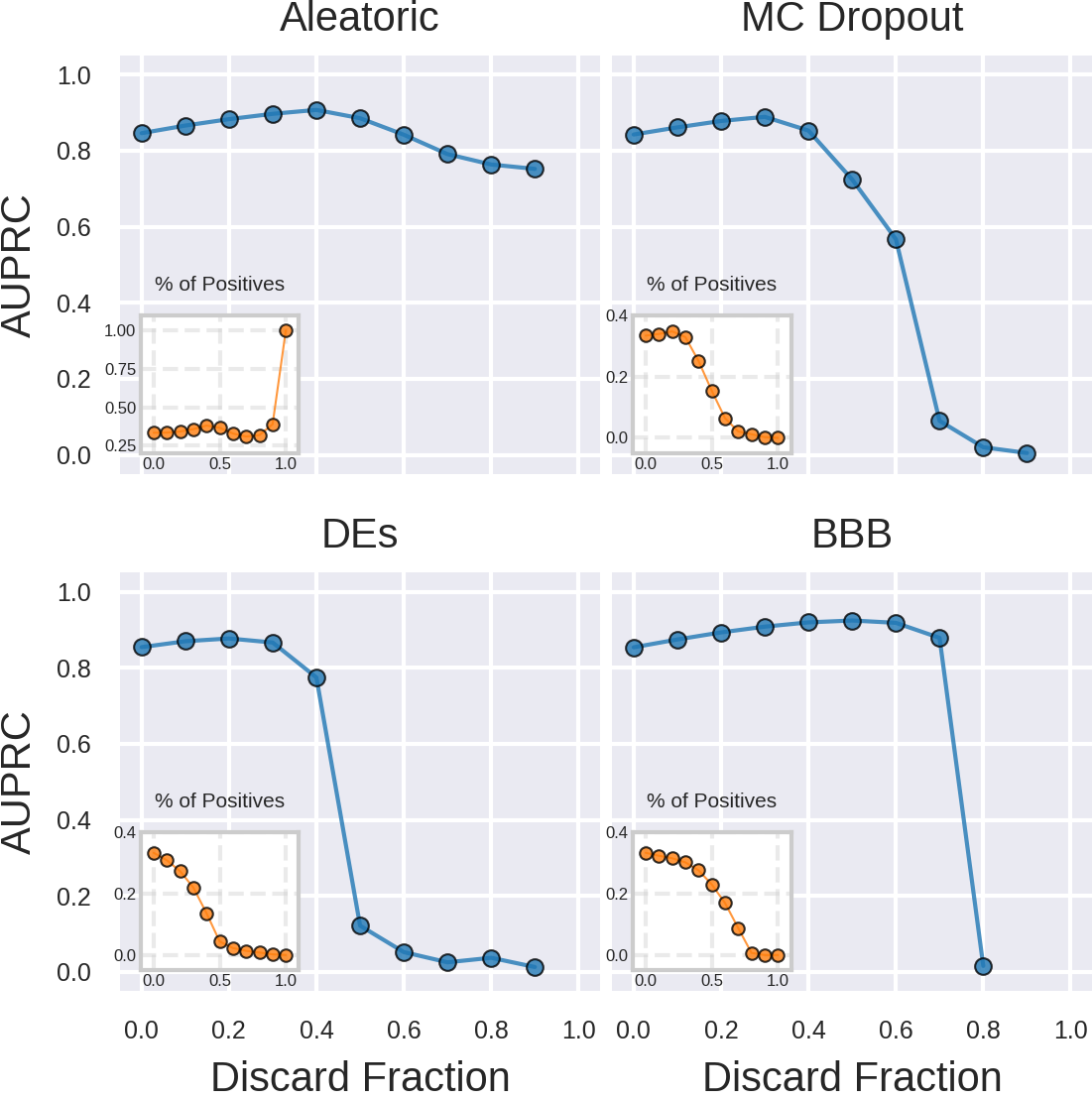}
    \caption{Discard test plots for all models trained with aleatoric uncertainty using the AUPRC as error measurement. A reliable model should exhibit increasing AUPRC score values as the discard fraction increases, indicating that keeping samples with higher uncertainty corresponds to lower AUPRC scores. The inset plots display the percentage of positive samples within each discard fraction.}
    \label{fig:sup_discard_auprc}
\end{figure}



\begin{figure}
  \centering
    \includegraphics[width=0.7\linewidth]{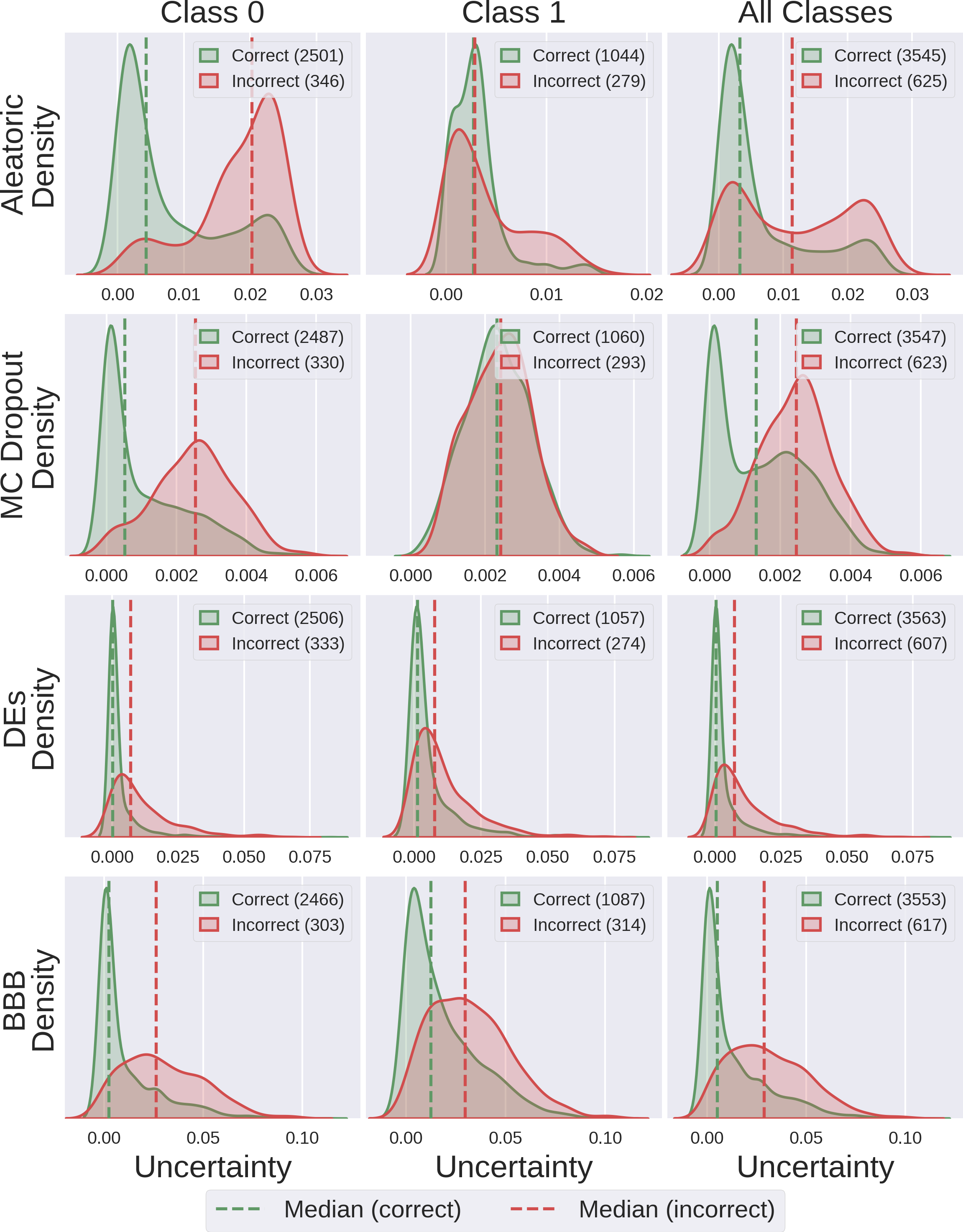}
    \caption{Uncertainty density plots across all models trained without aleatoric uncertainty. The plots are presented for the two classes combined as well as separately for the positive and negative classes. Vertical dashed lines mark the median uncertainty for each group. Reliable uncertainty estimates are characterized by clearly separated distributions with minimal overlap.}
    \label{fig:sup_density}
\end{figure}

%% file: bibliography.bib
@article{hodges_wildland_2019,
	title = {Wildland {Fire} {Spread} {Modeling} {Using} {Convolutional} {Neural} {Networks}},
	volume = {55},
	issn = {1572-8099},
	url = {https://doi.org/10.1007/s10694-019-00846-4},
	doi = {10.1007/s10694-019-00846-4},
	language = {en},
	number = {6},
	urldate = {2021-01-22},
	journal = {Fire Technology},
	author = {Hodges, Jonathan L. and Lattimer, Brian Y.},
	month = nov,
	year = {2019},
	pages = {2115--2142},
}

@misc{burge_convolutional_2021,
	title = {Convolutional {LSTM} {Neural} {Networks} for {Modeling} {Wildland} {Fire} {Dynamics}},
	url = {http://arxiv.org/abs/2012.06679},
	urldate = {2023-10-16},
	publisher = {arXiv},
	author = {Burge, John and Bonanni, Matthew and Ihme, Matthias and Hu, Lily},
	month = apr,
	year = {2021},
	note = {arXiv:2012.06679 [cs]},
}

@inproceedings{radke_firecast_2019,
	address = {Macao, China},
	title = {{FireCast}: {Leveraging} {Deep} {Learning} to {Predict} {Wildfire} {Spread}},
	isbn = {978-0-9992411-4-1},
	shorttitle = {{FireCast}},
	url = {https://www.ijcai.org/proceedings/2019/636},
	doi = {10.24963/ijcai.2019/636},
	language = {en},
	urldate = {2023-10-16},
	booktitle = {Proceedings of the {Twenty}-{Eighth} {International} {Joint} {Conference} on {Artificial} {Intelligence}},
	publisher = {International Joint Conferences on Artificial Intelligence Organization},
	author = {Radke, David and Hessler, Anna and Ellsworth, Dan},
	month = aug,
	year = {2019},
	pages = {4575--4581},
}

@article{zhao_tokenized_2023,
	title = {Tokenized {Time}-{Series} in {Satellite} {Image} {Segmentation} {With} {Transformer} {Network} for {Active} {Fire} {Detection}},
	volume = {61},
	issn = {0196-2892, 1558-0644},
	url = {https://ieeexplore.ieee.org/document/10155171/},
	doi = {10.1109/TGRS.2023.3287498},
	language = {en},
	urldate = {2023-11-07},
	journal = {IEEE Transactions on Geoscience and Remote Sensing},
	author = {Zhao, Yu and Ban, Yifang and Sullivan, Josephine},
	year = {2023},
	pages = {1--13},
}

@article{sdraka2024floga,
  title={FLOGA: A machine learning ready dataset, a benchmark and a novel deep learning model for burnt area mapping with Sentinel-2},
  author={Sdraka, Maria and Dimakos, Alkinoos and Malounis, Alexandros and Ntasiou, Zisoula and Karantzalos, Konstantinos and Michail, Dimitrios and Papoutsis, Ioannis},
  journal={IEEE Journal of Selected Topics in Applied Earth Observations and Remote Sensing},
  year={2024},
  publisher={IEEE}
}

@article{ghali2022deep,
	title = {Deep {Learning} and {Transformer} {Approaches} for {UAV}-{Based} {Wildfire} {Detection} and {Segmentation}},
	volume = {22},
	issn = {1424-8220},
	doi = {10.3390/s22051977},
	language = {eng},
	number = {5},
	journal = {Sensors (Basel, Switzerland)},
	author = {Ghali, Rafik and Akhloufi, Moulay A. and Mseddi, Wided Souidene},
	month = mar,
	year = {2022},
	pmid = {35271126},
	pmcid = {PMC8914964},
	pages = {1977},
}

@article{bouguettaya2022review,
title = {A review on early wildfire detection from unmanned aerial vehicles using deep learning-based computer vision algorithms},
journal = {Signal Processing},
volume = {190},
pages = {108309},
year = {2022},
issn = {0165-1684},
doi = {https://doi.org/10.1016/j.sigpro.2021.108309},
url = {https://www.sciencedirect.com/science/article/pii/S0165168421003467},
author = {Abdelmalek Bouguettaya and Hafed Zarzour and Amine Mohammed Taberkit and Ahmed Kechida}
}

@InProceedings{Prapas_2023_ICCV,
    author    = {Prapas, Ioannis and Bountos, Nikolaos-Ioannis and Kondylatos, Spyros and Michail, Dimitrios and Camps-Valls, Gustau and Papoutsis, Ioannis},
    title     = {TeleViT: Teleconnection-Driven Transformers Improve Subseasonal to Seasonal Wildfire Forecasting},
    booktitle = {Proceedings of the IEEE/CVF International Conference on Computer Vision (ICCV) Workshops},
    month     = {October},
    year      = {2023},
    pages     = {3754-3759}
}

@article{ghali_2023,
AUTHOR = {Ghali, Rafik and Akhloufi, Moulay A.},
TITLE = {Deep Learning Approaches for Wildland Fires Using Satellite Remote Sensing Data: Detection, Mapping, and Prediction},
JOURNAL = {Fire},
VOLUME = {6},
YEAR = {2023},
NUMBER = {5},
ARTICLE-NUMBER = {192},
URL = {https://www.mdpi.com/2571-6255/6/5/192},
ISSN = {2571-6255},
DOI = {10.3390/fire6050192}
}

@article{shadrin2024wildfire,
	title = {Wildfire spreading prediction using multimodal data and deep neural network approach},
	volume = {14},
	copyright = {2024 The Author(s)},
	issn = {2045-2322},
	url = {https://www.nature.com/articles/s41598-024-52821-x},
	doi = {10.1038/s41598-024-52821-x},
	language = {en},
	number = {1},
	urldate = {2025-07-24},
	journal = {Scientific Reports},
	author = {Shadrin, Dmitrii and Illarionova, Svetlana and Gubanov, Fedor and Evteeva, Ksenia and Mironenko, Maksim and Levchunets, Ivan and Belousov, Roman and Burnaev, Evgeny},
	month = jan,
	year = {2024},
	note = {Publisher: Nature Publishing Group},
	pages = {2606},
}

@inproceedings{stahl_evaluation_2020,
	address = {Cham},
	title = {Evaluation of {Uncertainty} {Quantification} in {Deep} {Learning}},
	isbn = {978-3-030-50146-4},
	doi = {10.1007/978-3-030-50146-4_41},
	language = {en},
	booktitle = {Information {Processing} and {Management} of {Uncertainty} in {Knowledge}-{Based} {Systems}},
	publisher = {Springer International Publishing},
	author = {Ståhl, Niclas and Falkman, Göran and Karlsson, Alexander and Mathiason, Gunnar},
	editor = {Lesot, Marie-Jeanne and Vieira, Susana and Reformat, Marek Z. and Carvalho, João Paulo and Wilbik, Anna and Bouchon-Meunier, Bernadette and Yager, Ronald R.},
	year = {2020},
	pages = {556--568},
}

@inproceedings{guo_calibration,
author = {Guo, Chuan and Pleiss, Geoff and Sun, Yu and Weinberger, Kilian Q.},
title = {On calibration of modern neural networks},
year = {2017},
publisher = {JMLR.org},
booktitle = {Proceedings of the 34th International Conference on Machine Learning - Volume 70},
pages = {1321–1330},
numpages = {10},
location = {Sydney, NSW, Australia},
series = {ICML'17}
}

@inproceedings{muscanyi_neurips_2024,
 author = {Mucs\'{a}nyi, B\'{a}lint and Kirchhof, Michael and Oh, Seong Joon},
 booktitle = {Advances in Neural Information Processing Systems},
 editor = {A. Globerson and L. Mackey and D. Belgrave and A. Fan and U. Paquet and J. Tomczak and C. Zhang},
 pages = {50972--51038},
 publisher = {Curran Associates, Inc.},
 title = {Benchmarking Uncertainty Disentanglement: Specialized Uncertainties for Specialized Tasks},
 url = {https://proceedings.neurips.cc/paper_files/paper/2024/file/5afa9cb1e917b898ad418216dc726fbd-Paper-Datasets_and_Benchmarks_Track.pdf},
 volume = {37},
 year = {2024}
}

@InProceedings{mukhoti_2023_cvpr,
    author    = {Mukhoti, Jishnu and Kirsch, Andreas and van Amersfoort, Joost and Torr, Philip H.S. and Gal, Yarin},
    title     = {Deep Deterministic Uncertainty: A New Simple Baseline},
    booktitle = {Proceedings of the IEEE/CVF Conference on Computer Vision and Pattern Recognition (CVPR)},
    month     = {June},
    year      = {2023},
    pages     = {24384-24394}
}

@article{deste_modeling_2020,
	title = {Modeling fire ignition probability and frequency using {Hurdle} models: a cross-regional study in {Southern} {Europe}},
	volume = {9},
	issn = {2192-1709},
	shorttitle = {Modeling fire ignition probability and frequency using {Hurdle} models},
	url = {https://doi.org/10.1186/s13717-020-00263-4},
	doi = {10.1186/s13717-020-00263-4},
	language = {en},
	number = {1},
	urldate = {2025-06-29},
	journal = {Ecological Processes},
	author = {D’Este, Marina and Ganga, Antonio and Elia, Mario and Lovreglio, Raffaella and Giannico, Vincenzo and Spano, Giuseppina and Colangelo, Giuseppe and Lafortezza, Raffaele and Sanesi, Giovanni},
	month = oct,
	year = {2020},
	pages = {54},
}

@article{elia_modeling_2019,
	title = {Modeling fire ignition patterns in {Mediterranean} urban interfaces},
	volume = {33},
	issn = {1436-3259},
	url = {https://doi.org/10.1007/s00477-018-1558-5},
	doi = {10.1007/s00477-018-1558-5},
	language = {en},
	number = {1},
	urldate = {2025-06-29},
	journal = {Stochastic Environmental Research and Risk Assessment},
	author = {Elia, M. and Giannico, V. and Lafortezza, R. and Sanesi, G.},
	month = jan,
	year = {2019},
	pages = {169--181},
}

@article{begueria_validation_2006,
	title = {Validation and {Evaluation} of {Predictive} {Models} in {Hazard} {Assessment} and {Risk} {Management}},
	volume = {37},
	issn = {1573-0840},
	url = {https://doi.org/10.1007/s11069-005-5182-6},
	doi = {10.1007/s11069-005-5182-6},
	language = {en},
	number = {3},
	urldate = {2025-07-01},
	journal = {Natural Hazards},
	author = {Beguería, Santiago},
	month = mar,
	year = {2006},
	pages = {315--329},
}

@inproceedings{ovadia_can_2019,
	title = {Can you trust your model' s uncertainty? {Evaluating} predictive uncertainty under dataset shift},
	volume = {32},
	shorttitle = {Can you trust your model' s uncertainty?},
	url = {https://proceedings.neurips.cc/paper/2019/hash/8558cb408c1d76621371888657d2eb1d-Abstract.html},
	urldate = {2024-05-02},
	booktitle = {Advances in {Neural} {Information} {Processing} {Systems}},
	publisher = {Curran Associates, Inc.},
	author = {Ovadia, Yaniv and Fertig, Emily and Ren, Jie and Nado, Zachary and Sculley, D. and Nowozin, Sebastian and Dillon, Joshua and Lakshminarayanan, Balaji and Snoek, Jasper},
	year = {2019},
}

@article{guikema_artificial_2020,
	title = {Artificial {Intelligence} for {Natural} {Hazards} {Risk} {Analysis}: {Potential}, {Challenges}, and {Research} {Needs}},
	volume = {40},
	copyright = {© 2020 Society for Risk Analysis},
	issn = {1539-6924},
	shorttitle = {Artificial {Intelligence} for {Natural} {Hazards} {Risk} {Analysis}},
	url = {https://onlinelibrary.wiley.com/doi/abs/10.1111/risa.13476},
	doi = {10.1111/risa.13476},
	language = {en},
	number = {6},
	urldate = {2025-07-01},
	journal = {Risk Analysis},
	author = {Guikema, Seth},
	year = {2020},
	note = {\_eprint: https://onlinelibrary.wiley.com/doi/pdf/10.1111/risa.13476},
	pages = {1117--1123},
}

@inproceedings{gustafsson_evaluating_2020,
	title = {Evaluating {Scalable} {Bayesian} {Deep} {Learning} {Methods} for {Robust} {Computer} {Vision}},
	url = {https://ieeexplore.ieee.org/document/9150658/},
	doi = {10.1109/CVPRW50498.2020.00167},
	urldate = {2025-07-01},
	booktitle = {2020 {IEEE}/{CVF} {Conference} on {Computer} {Vision} and {Pattern} {Recognition} {Workshops} ({CVPRW})},
	author = {Gustafsson, Fredrik K. and Danelljan, Martin and Schon, Thomas B.},
	month = jun,
	year = {2020},
	note = {ISSN: 2160-7516},
	pages = {1289--1298},
}

@inproceedings{
kirchhof2023url,
title={{URL}: A Representation Learning Benchmark for Transferable Uncertainty Estimates},
author={Michael Kirchhof and B{\'a}lint Mucs{\'a}nyi and Seong Joon Oh and Enkelejda Kasneci},
booktitle={Thirty-seventh Conference on Neural Information Processing Systems Datasets and Benchmarks Track},
year={2023},
url={https://openreview.net/forum?id=e9n4JjkmXZ}
}

@ARTICLE{persello_2021,
  author={Persello, Claudio and Wegner, Jan Dirk and Hänsch, Ronny and Tuia, Devis and Ghamisi, Pedram and Koeva, Mila and Camps-Valls, Gustau},
  journal={IEEE Geoscience and Remote Sensing Magazine}, 
  title={Deep Learning and Earth Observation to Support the Sustainable Development Goals: Current approaches, open challenges, and future opportunities}, 
  year={2022},
  volume={10},
  number={2},
  pages={172-200},
  doi={10.1109/MGRS.2021.3136100}}

@Article{hantson_2016,
AUTHOR = {Hantson, S. and Arneth, A. and Harrison, S. P. and Kelley, D. I. and Prentice, I. C. and Rabin, S. S. and Archibald, S. and Mouillot, F. and Arnold, S. R. and Artaxo, P. and Bachelet, D. and Ciais, P. and Forrest, M. and Friedlingstein, P. and Hickler, T. and Kaplan, J. O. and Kloster, S. and Knorr, W. and Lasslop, G. and Li, F. and Mangeon, S. and Melton, J. R. and Meyn, A. and Sitch, S. and Spessa, A. and van der Werf, G. R. and Voulgarakis, A. and Yue, C.},
TITLE = {The status and challenge of global fire modelling},
JOURNAL = {Biogeosciences},
VOLUME = {13},
YEAR = {2016},
NUMBER = {11},
PAGES = {3359--3375},
URL = {https://bg.copernicus.org/articles/13/3359/2016/},
DOI = {10.5194/bg-13-3359-2016}
}

@article{zhang_calibration_2025,
	title = {Calibration and uncertainty quantification for deep learning-based drought detection},
	volume = {140},
	issn = {1569-8432},
	url = {https://www.sciencedirect.com/science/article/pii/S1569843225002109},
	doi = {10.1016/j.jag.2025.104563},
	urldate = {2025-06-16},
	journal = {International Journal of Applied Earth Observation and Geoinformation},
	author = {Zhang, Mengxue and Fernández-Torres, Miguel-\'Angel and Cohrs, Kai-Hendrik and Camps-Valls, Gustau},
	month = jun,
	year = {2025},
	pages = {104563},
}

@article{haynes_creating_2023,
	title = {Creating and {Evaluating} {Uncertainty} {Estimates} with {Neural} {Networks} for {Environmental}-{Science} {Applications}},
	volume = {2},
	issn = {2769-7525},
	url = {https://journals.ametsoc.org/view/journals/aies/2/2/AIES-D-22-0061.1.xml},
	doi = {10.1175/AIES-D-22-0061.1},
	language = {EN},
	number = {2},
	urldate = {2023-04-25},
	journal = {Artificial Intelligence for the Earth Systems},
	author = {Haynes, Katherine and Lagerquist, Ryan and McGraw, Marie and Musgrave, Kate and Ebert-Uphoff, Imme},
	month = apr,
	year = {2023},
	note = {Publisher: American Meteorological Society
Section: Artificial Intelligence for the Earth Systems},
}

@inproceedings{mesogeos_kondylatos,
 author = {Kondylatos, Spyridon and Prapas, Ioannis and Camps-Valls, Gustau and Papoutsis, Ioannis},
 booktitle = {Advances in Neural Information Processing Systems},
 editor = {A. Oh and T. Naumann and A. Globerson and K. Saenko and M. Hardt and S. Levine},
 pages = {50661--50676},
 publisher = {Curran Associates, Inc.},
 title = {Mesogeos: A multi-purpose dataset for data-driven wildfire modeling in the Mediterranean},
 url = {https://proceedings.neurips.cc/paper_files/paper/2023/file/9ee3ed2dd656402f954ef9dc37e39f48-Paper-Datasets_and_Benchmarks.pdf},
 volume = {36},
 year = {2023}
}

@article{huot2022next,
  title={Next day wildfire spread: A machine learning dataset to predict wildfire spreading from remote-sensing data},
  author={Huot, Fantine and Hu, R Lily and Goyal, Nita and Sankar, Tharun and Ihme, Matthias and Chen, Yi-Fan},
  journal={IEEE Transactions on Geoscience and Remote Sensing},
  volume={60},
  pages={1--13},
  year={2022},
  publisher={IEEE}
}

@inproceedings{NEURIPS2023_ebd54517,
 author = {Gerard, Sebastian and Zhao, Yu and Sullivan, Josephine},
 booktitle = {Advances in Neural Information Processing Systems},
 editor = {A. Oh and T. Naumann and A. Globerson and K. Saenko and M. Hardt and S. Levine},
 pages = {74515--74529},
 publisher = {Curran Associates, Inc.},
 title = {WildfireSpreadTS: A dataset of multi-modal time series for wildfire spread prediction},
 volume = {36},
 year = {2023}
}

@Article{knopp2020deep,
AUTHOR = {Knopp, Lisa and Wieland, Marc and Rättich, Michaela and Martinis, Sandro},
TITLE = {A Deep Learning Approach for Burned Area Segmentation with Sentinel-2 Data},
JOURNAL = {Remote Sensing},
VOLUME = {12},
YEAR = {2020},
NUMBER = {15},
ARTICLE-NUMBER = {2422},
URL = {https://www.mdpi.com/2072-4292/12/15/2422},
ISSN = {2072-4292},
DOI = {10.3390/rs12152422}
}

@article{guikema_natural_hazards,
author = {Guikema, Seth},
title = {Artificial Intelligence for Natural Hazards Risk Analysis: Potential, Challenges, and Research Needs},
journal = {Risk Analysis},
volume = {40},
number = {6},
pages = {1117-1123},
doi = {https://doi.org/10.1111/risa.13476},
url = {https://onlinelibrary.wiley.com/doi/abs/10.1111/risa.13476},
eprint = {https://onlinelibrary.wiley.com/doi/pdf/10.1111/risa.13476},
year = {2020}
}

@article{sun_applications_2020,
	title = {Applications of artificial intelligence for disaster management},
	volume = {103},
	issn = {1573-0840},
	url = {https://doi.org/10.1007/s11069-020-04124-3},
	doi = {10.1007/s11069-020-04124-3},
	number = {3},
	urldate = {2025-05-05},
	journal = {Natural Hazards},
	author = {Sun, Wenjuan and Bocchini, Paolo and Davison, Brian D.},
	month = sep,
	year = {2020},
	pages = {2631--2689},
}

@article{HIMEUR202244,
title = {Using artificial intelligence and data fusion for environmental monitoring: A review and future perspectives},
journal = {Information Fusion},
volume = {86-87},
pages = {44-75},
year = {2022},
issn = {1566-2535},
doi = {https://doi.org/10.1016/j.inffus.2022.06.003},
url = {https://www.sciencedirect.com/science/article/pii/S1566253522000574},
author = {Yassine Himeur and Bhagawat Rimal and Abhishek Tiwary and Abbes Amira}
}

@article{bostrom_trust,
	title = {Trust and trustworthy artificial intelligence: {A} research agenda for {AI} in the environmental sciences},
	shorttitle = {Trust and trustworthy artificial intelligence},
	url = {https://opensky.ucar.edu/islandora/object/%3A21428},
    doi = { https://doi.org/10.1111/risa.14245},
	language = {en},
	urldate = {2025-05-05},
	journal = {Risk Analysis},
    year = {2024},
	author = {Bostrom, Author: A. and Demuth, Author: Julie L. and Wirz, Author: Christopher D. and Cains, Author: Mariana and Schumacher, Author: Andrea and Madlambayan, Author: D. and Bansal, Author: A. S. and Bearth, Author: A. and Chase, Author: R. and Crosman, Author: K. M. and Ebert-Uphoff, Author: I. and Gagne, Author: David John and Guikema, Author: S. and Hoffman, Author: R. and Johnson, Author: B. B. and Kumler-Bonfanti, Author: C. and Lee, Author: J. D. and Lowe, Author: A. and McGovern, Author: A. and Przybylo, Author: V. and Radford, Author: J. T. and Roth, Author: E. and Sutter, Author: C. and Tissot, Author: P. and Roebber, Author: P. and Stewart, Author: J. Q. and White, Author: M. and Williams, Author: J. K.},
}

@article{salcedo_analysis,
title = "Analysis, characterization, prediction, and attribution of extreme atmospheric events with machine learning and deep learning techniques: a review",
author = "Sancho Salcedo-Sanz and Jorge P{\'e}rez-Aracil and Guido Ascenso and {Del Ser}, Javier and David Casillas-P{\'e}rez and Christopher Kadow and Du{\v s}an Fister and David Barriopedro and Ricardo Garc{\'i}a-Herrera and Matteo Giuliani and Andrea Castelletti",
note = "Publisher Copyright: {\textcopyright} 2023, The Author(s).",
year = "2024",
month = jan,
doi = "10.1007/s00704-023-04571-5",
language = "English",
volume = "155",
pages = "1--44",
journal = "Theoretical and Applied Climatology",
issn = "0177-798X",
publisher = "Springer-Verlag Wien",
number = "1",
}

@article{camps-valls_artificial_2025,
	title = {Artificial intelligence for modeling and understanding extreme weather and climate events},
	volume = {16},
	copyright = {2025 The Author(s)},
	issn = {2041-1723},
	url = {https://www.nature.com/articles/s41467-025-56573-8},
	doi = {10.1038/s41467-025-56573-8},
	language = {en},
	number = {1},
	urldate = {2025-05-05},
	journal = {Nature Communications},
	author = {Camps-Valls, Gustau and Fernández-Torres, Miguel-\'Angel and Cohrs, Kai-Hendrik and Höhl, Adrian and Castelletti, Andrea and Pacal, Aytac and Robin, Claire and Martinuzzi, Francesco and Papoutsis, Ioannis and Prapas, Ioannis and Pérez-Aracil, Jorge and Weigel, Katja and Gonzalez-Calabuig, Maria and Reichstein, Markus and Rabel, Martin and Giuliani, Matteo and Mahecha, Miguel D. and Popescu, Oana-Iuliana and Pellicer-Valero, Oscar J. and Ouala, Said and Salcedo-Sanz, Sancho and Sippel, Sebastian and Kondylatos, Spyros and Happé, Tamara and Williams, Tristan},
	month = feb,
	year = {2025},
	note = {Publisher: Nature Publishing Group},
	pages = {1919},
}

@article{albahri_systematic_2024,
	title = {A systematic review of trustworthy artificial intelligence applications in natural disasters},
	volume = {118},
	issn = {0045-7906},
	url = {https://www.sciencedirect.com/science/article/pii/S0045790624003379},
	doi = {10.1016/j.compeleceng.2024.109409},
	urldate = {2025-05-06},
	journal = {Computers and Electrical Engineering},
	author = {Albahri, A. S. and Khaleel, Yahya Layth and Habeeb, Mustafa Abdulfattah and Ismael, Reem D. and Hameed, Qabas A. and Deveci, Muhammet and Homod, Raad Z. and Albahri, O. S. and Alamoodi, A. H. and Alzubaidi, Laith},
	month = sep,
	year = {2024},
	pages = {109409},
}

@article{moreira_wildfire_2020,
	title = {Wildfire management in {Mediterranean}-type regions: paradigm change needed},
	volume = {15},
	issn = {1748-9326},
	shorttitle = {Wildfire management in {Mediterranean}-type regions},
	url = {https://iopscience.iop.org/article/10.1088/1748-9326/ab541e},
	doi = {10.1088/1748-9326/ab541e},
	language = {en},
	number = {1},
	urldate = {2022-03-29},
	journal = {Environmental Research Letters},
	author = {Moreira, Francisco and Ascoli, Davide and Safford, Hugh and Adams, Mark A and Moreno, José M and Pereira, José M C and Catry, Filipe X and Armesto, Juan and Bond, William and González, Mauro E and Curt, Thomas and Koutsias, Nikos and McCaw, Lachlan and Price, Owen and Pausas, Juli G and Rigolot, Eric and Stephens, Scott and Tavsanoglu, Cagatay and Vallejo, V Ramon and Van Wilgen, Brian W and Xanthopoulos, Gavriil and Fernandes, Paulo M},
	month = jan,
	year = {2020},
	pages = {011001},
}

@article{ruffault_increased_2020,
	title = {Increased likelihood of heat-induced large wildfires in the {Mediterranean} {Basin}},
	volume = {10},
	issn = {2045-2322},
	url = {https://www.nature.com/articles/s41598-020-70069-z},
	doi = {10.1038/s41598-020-70069-z},
	language = {en},
	number = {1},
	urldate = {2022-03-23},
	journal = {Scientific Reports},
	author = {Ruffault, Julien and Curt, Thomas and Moron, Vincent and Trigo, Ricardo M. and Mouillot, Florent and Koutsias, Nikos and Pimont, François and Martin-StPaul, Nicolas and Barbero, Renaud and Dupuy, Jean-Luc and Russo, Ana and Belhadj-Khedher, Chiraz},
	month = dec,
	year = {2020},
	pages = {13790},
}

@misc{kondylatos_probabilistic_2025,
	title = {Probabilistic {Machine} {Learning} for {Noisy} {Labels} in {Earth} {Observation}},
	url = {http://arxiv.org/abs/2504.03478},
	doi = {10.48550/arXiv.2504.03478},
	urldate = {2025-05-05},
	publisher = {arXiv},
	author = {Kondylatos, Spyros and Bountos, Nikolaos Ioannis and Prapas, Ioannis and Zavras, Angelos and Camps-Valls, Gustau and Papoutsis, Ioannis},
	month = apr,
	year = {2025},
	note = {arXiv:2504.03478 [cs]},
}

@article{huot_deep_2020,
	title = {Deep {Learning} {Models} for {Predicting} {Wildfires} from {Historical} {Remote}-{Sensing} {Data}},
	url = {http://arxiv.org/abs/2010.07445},
	urldate = {2020-12-23},
	journal = {arXiv:2010.07445 [cs]},
	author = {Huot, Fantine and Hu, R. Lily and Ihme, Matthias and Wang, Qing and Burge, John and Lu, Tianjian and Hickey, Jason and Chen, Yi-Fan and Anderson, John},
	month = nov,
	year = {2020},
	note = {arXiv: 2010.07445},
}

@article{zhang_deep_2021,
	title = {Deep neural networks for global wildfire susceptibility modelling},
	volume = {127},
	issn = {1470-160X},
	url = {https://www.sciencedirect.com/science/article/pii/S1470160X21004003},
	doi = {10.1016/j.ecolind.2021.107735},
	language = {en},
	urldate = {2021-09-10},
	journal = {Ecological Indicators},
	author = {Zhang, Guoli and Wang, Ming and Liu, Kai},
	month = aug,
	year = {2021},
	pages = {107735},
}

@article{bergado_predicting_2021,
	title = {Predicting wildfire burns from big geodata using deep learning},
	volume = {140},
	issn = {0925-7535},
	url = {https://www.sciencedirect.com/science/article/pii/S0925753521001211},
	doi = {10.1016/j.ssci.2021.105276},
	language = {en},
	urldate = {2021-09-10},
	journal = {Safety Science},
	author = {Bergado, John Ray and Persello, Claudio and Reinke, Karin and Stein, Alfred},
	month = aug,
	year = {2021},
	pages = {105276},
}

@article{bjanes_deep_2021,
	title = {A deep learning ensemble model for wildfire susceptibility mapping},
	volume = {65},
	issn = {1574-9541},
	url = {https://www.sciencedirect.com/science/article/pii/S1574954121001886},
	doi = {10.1016/j.ecoinf.2021.101397},
	language = {en},
	urldate = {2022-02-28},
	journal = {Ecological Informatics},
	author = {Bjånes, Alexandra and De La Fuente, Rodrigo and Mena, Pablo},
	month = aug,
	year = {2021},
	pages = {101397},
}

@article{zhang_forest_2019,
	title = {Forest {Fire} {Susceptibility} {Modeling} {Using} a {Convolutional} {Neural} {Network} for {Yunnan} {Province} of {China}},
	volume = {10},
	issn = {2095-0055, 2192-6395},
	url = {http://link.springer.com/10.1007/s13753-019-00233-1},
	doi = {10.1007/s13753-019-00233-1},
	language = {en},
	number = {3},
	urldate = {2022-06-21},
	journal = {International Journal of Disaster Risk Science},
	author = {Zhang, Guoli and Wang, Ming and Liu, Kai},
	month = sep,
	year = {2019},
	pages = {386--403},
}

@article{kondylatos_wildfire_2022,
	title = {Wildfire {Danger} {Prediction} and {Understanding} {With} {Deep} {Learning}},
	volume = {49},
	issn = {1944-8007},
	url = {https://onlinelibrary.wiley.com/doi/abs/10.1029/2022GL099368},
	doi = {10.1029/2022GL099368},
	language = {en},
	number = {17},
	urldate = {2022-12-21},
	journal = {Geophysical Research Letters},
	author = {Kondylatos, Spyros and Prapas, Ioannis and Ronco, Michele and Papoutsis, Ioannis and Camps-Valls, Gustau and Piles, María and Fernández-Torres, Miguel-\'Angel and Carvalhais, Nuno},
	year = {2022},
	note = {\_eprint: https://onlinelibrary.wiley.com/doi/pdf/10.1029/2022GL099368},
	pages = {e2022GL099368},
}

@article{mousavi_bayesian-deep-learning_2020,
	title = {Bayesian-{Deep}-{Learning} {Estimation} of {Earthquake} {Location} from {Single}-{Station} {Observations}},
	volume = {58},
	issn = {0196-2892, 1558-0644},
	url = {http://arxiv.org/abs/1912.01144},
	doi = {10.1109/TGRS.2020.2988770},
	language = {en},
	number = {11},
	urldate = {2025-05-07},
	journal = {IEEE Transactions on Geoscience and Remote Sensing},
	author = {Mousavi, S. Mostafa and Beroza, Gregory C.},
	month = nov,
	year = {2020},
	note = {arXiv:1912.01144 [physics]},
	pages = {8211--8224},
}

@article{bueno_volcano-seismic_2020,
	title = {Volcano-{Seismic} {Transfer} {Learning} and {Uncertainty} {Quantification} {With} {Bayesian} {Neural} {Networks}},
	volume = {58},
	issn = {1558-0644},
	url = {https://ieeexplore.ieee.org/abstract/document/8861294},
	doi = {10.1109/TGRS.2019.2941494},
	urldate = {2025-05-07},
	journal = {IEEE Transactions on Geoscience and Remote Sensing},
	author = {Bueno, Angel and Benítez, Carmen and De Angelis, Silvio and Díaz Moreno, Alejandro and Ibáñez, Jesús M.},
	month = feb,
	year = {2020},
	pages = {892--902},
}

@article{gamboa-chacon_analysis_2025,
	title = {Analysis of earthquake detection using deep learning: {Evaluating} reliability and uncertainty in prediction methods},
	volume = {197},
	issn = {0098-3004},
	shorttitle = {Analysis of earthquake detection using deep learning},
	url = {https://www.sciencedirect.com/science/article/pii/S0098300425000275},
	doi = {10.1016/j.cageo.2025.105877},
	journal = {Computers \& Geosciences},
	author = {Gamboa-Chacón, Sebastián and Meneses, Esteban and Chaves, Esteban J.},
	month = mar,
	year = {2025},
	pages = {105877},
}

@inproceedings{wang_deep_2019,
	address = {New York, NY, USA},
	series = {{KDD} '19},
	title = {Deep {Uncertainty} {Quantification}: {A} {Machine} {Learning} {Approach} for {Weather} {Forecasting}},
	isbn = {978-1-4503-6201-6},
	shorttitle = {Deep {Uncertainty} {Quantification}},
	url = {https://dl.acm.org/doi/10.1145/3292500.3330704},
	doi = {10.1145/3292500.3330704},
	urldate = {2025-05-07},
	booktitle = {Proceedings of the 25th {ACM} {SIGKDD} {International} {Conference} on {Knowledge} {Discovery} \& {Data} {Mining}},
	publisher = {Association for Computing Machinery},
	author = {Wang, Bin and Lu, Jie and Yan, Zheng and Luo, Huaishao and Li, Tianrui and Zheng, Yu and Zhang, Guangquan},
	month = jul,
	year = {2019},
	pages = {2087--2095},
}

@article{ferchichi_evidential_2025,
	title = {Evidential uncertainty quantification with multiple deep learning architectures for spatiotemporal drought forecasting},
	volume = {37},
	issn = {1433-3058},
	url = {https://doi.org/10.1007/s00521-025-11026-7},
	doi = {10.1007/s00521-025-11026-7},
	language = {en},
	number = {15},
	urldate = {2025-05-07},
	journal = {Neural Computing and Applications},
	author = {Ferchichi, Ahlem and Chihaoui, Mejda and Toujani, Radhia and Ferchichi, Aya and Hendaoui, Fatma},
	month = may,
	year = {2025},
	pages = {8773--8797},
}

@article{klotz_uncertainty_2022,
	title = {Uncertainty estimation with deep learning for rainfall–runoff modeling},
	volume = {26},
	issn = {1027-5606},
	url = {https://hess.copernicus.org/articles/26/1673/2022/},
	doi = {10.5194/hess-26-1673-2022},
	language = {English},
	number = {6},
	urldate = {2025-05-07},
	journal = {Hydrology and Earth System Sciences},
	author = {Klotz, Daniel and Kratzert, Frederik and Gauch, Martin and Keefe Sampson, Alden and Brandstetter, Johannes and Klambauer, Günter and Hochreiter, Sepp and Nearing, Grey},
	month = mar,
	year = {2022},
	note = {Publisher: Copernicus GmbH},
	pages = {1673--1693},
}

@article{wang_quantification_2023,
	series = {Data driven models},
	title = {Quantification of model uncertainty and variability for landslide displacement prediction based on {Monte} {Carlo} simulation},
	volume = {123},
	issn = {1342-937X},
	url = {https://www.sciencedirect.com/science/article/pii/S1342937X23000801},
	doi = {10.1016/j.gr.2023.03.006},
	urldate = {2025-05-07},
	journal = {Gondwana Research},
	author = {Wang, Luqi and Xiao, Ting and Liu, Songlin and Zhang, Wengang and Yang, Beibei and Chen, Lichuan},
	month = nov,
	year = {2023},
	pages = {27--40},
}

@article{KALANTARI2019393,
title = {Assessing flood probability for transportation infrastructure based on catchment characteristics, sediment connectivity and remotely sensed soil moisture},
journal = {Science of The Total Environment},
volume = {661},
pages = {393-406},
year = {2019},
issn = {0048-9697},
doi = {https://doi.org/10.1016/j.scitotenv.2019.01.009},
url = {https://www.sciencedirect.com/science/article/pii/S0048969719300099},
author = {Zahra Kalantari and Carla Sofia Santos Ferreira and Alexander J. Koutsouris and Anna-Klara Ahlmer and Artemi Cerdà and Georgia Destouni}
}

@article{pausas_burning_2009,
	title = {A {Burning} {Story}: {The} {Role} of {Fire} in the {History} of {Life}},
	volume = {59},
	issn = {0006-3568, 1525-3244},
	shorttitle = {A {Burning} {Story}},
	url = {https://academic.oup.com/bioscience/article-lookup/doi/10.1525/bio.2009.59.7.10},
	doi = {10.1525/bio.2009.59.7.10},
	language = {en},
	number = {7},
	urldate = {2022-03-31},
	journal = {BioScience},
	author = {Pausas, Juli G. and Keeley, Jon E.},
	month = jul,
	year = {2009},
	pages = {593--601},
}

@article{pettinari_fire_2020,
	title = {Fire {Danger} {Observed} from {Space}},
	volume = {41},
	issn = {1573-0956},
	url = {https://doi.org/10.1007/s10712-020-09610-8},
	doi = {10.1007/s10712-020-09610-8},
	language = {en},
	number = {6},
	urldate = {2021-01-03},
	journal = {Surveys in Geophysics},
	author = {Pettinari, M. Lucrecia and Chuvieco, Emilio},
	month = nov,
	year = {2020},
	pages = {1437--1459},
}

@article{pausas_wildfires_2021,
	title = {Wildfires and global change},
	volume = {19},
	issn = {1540-9309},
	url = {https://onlinelibrary.wiley.com/doi/abs/10.1002/fee.2359},
	doi = {10.1002/fee.2359},
	language = {en},
	number = {7},
	urldate = {2022-03-30},
	journal = {Frontiers in Ecology and the Environment},
	author = {Pausas, Juli G and Keeley, Jon E},
	year = {2021},
	note = {\_eprint: https://onlinelibrary.wiley.com/doi/pdf/10.1002/fee.2359},
	pages = {387--395},
}

@article{archibald_defining,
author = {Sally Archibald  and Caroline E. R. Lehmann  and Jose L. Gómez-Dans  and Ross A. Bradstock },
title = {Defining pyromes and global syndromes of fire regimes},
journal = {Proceedings of the National Academy of Sciences},
volume = {110},
number = {16},
pages = {6442-6447},
year = {2013},
doi = {10.1073/pnas.1211466110},
URL = {https://www.pnas.org/doi/abs/10.1073/pnas.1211466110},
eprint = {https://www.pnas.org/doi/pdf/10.1073/pnas.1211466110}}

@article{jain_review_2020,
author = {Jain, Piyush and Coogan, Sean C.P. and Subramanian, Sriram Ganapathi and Crowley, Mark and Taylor, Steve and Flannigan, Mike D.},
title = {A review of machine learning applications in wildfire science and management},
journal = {Environmental Reviews},
volume = {28},
number = {4},
pages = {478-505},
year = {2020},
doi = {10.1139/er-2020-0019},
URL = {https://doi.org/10.1139/er-2020-0019},
eprint = {https://doi.org/10.1139/er-2020-0019}
}

@article{reichstein_deep_2019,
	title = {Deep learning and process understanding for data-driven {Earth} system science},
	volume = {566},
	number = {7743},
	journal = {Nature},
	author = {Reichstein, Markus and Camps-Valls, Gustau and Stevens, Bjorn and Jung, Martin and Denzler, Joachim and Carvalhais, Nuno},
	year = {2019},
	note = {Publisher: Nature Publishing Group},
	pages = {195--204},
}

@article{soille_versatile_2018,
	title = {A versatile data-intensive computing platform for information retrieval from big geospatial data},
	volume = {81},
	issn = {0167-739X},
	url = {https://www.sciencedirect.com/science/article/pii/S0167739X1730078X},
	doi = {10.1016/j.future.2017.11.007},
	urldate = {2023-09-25},
	journal = {Future Generation Computer Systems},
	author = {Soille, P. and Burger, A. and De Marchi, D. and Kempeneers, P. and Rodriguez, D. and Syrris, V. and Vasilev, V.},
	month = apr,
	year = {2018},
	pages = {30--40},
}

@article{srivastava_dropout,
  author  = {Nitish Srivastava and Geoffrey Hinton and Alex Krizhevsky and Ilya Sutskever and Ruslan Salakhutdinov},
  title   = {Dropout: A Simple Way to Prevent Neural Networks from Overfitting},
  journal = {Journal of Machine Learning Research},
  year    = {2014},
  volume  = {15},
  number  = {56},
  pages   = {1929--1958},
  url     = {http://jmlr.org/papers/v15/srivastava14a.html}
}

@inproceedings{malinin_predictive_2018,
author = {Malinin, Andrey and Gales, Mark},
title = {Predictive uncertainty estimation via prior networks},
year = {2018},
publisher = {Curran Associates Inc.},
address = {Red Hook, NY, USA},
booktitle = {Proceedings of the 32nd International Conference on Neural Information Processing Systems},
pages = {7047–7058},
numpages = {12},
location = {Montr\'{e}al, Canada},
series = {NIPS'18}
}

@inproceedings{sensoy_evidential_2018,
author = {Sensoy, Murat and Kaplan, Lance and Kandemir, Melih},
title = {Evidential deep learning to quantify classification uncertainty},
year = {2018},
publisher = {Curran Associates Inc.},
address = {Red Hook, NY, USA},
booktitle = {Proceedings of the 32nd International Conference on Neural Information Processing Systems},
pages = {3183–3193},
numpages = {11},
location = {Montr\'{e}al, Canada},
series = {NIPS'18}
}

@inproceedings{lee_gradients_2020,
	title = {Gradients as a {Measure} of {Uncertainty} in {Neural} {Networks}},
	url = {https://ieeexplore.ieee.org/document/9190679/},
	doi = {10.1109/ICIP40778.2020.9190679},
	urldate = {2025-07-24},
	booktitle = {2020 {IEEE} {International} {Conference} on {Image} {Processing} ({ICIP})},
	author = {Lee, Jinsol and AlRegib, Ghassan},
	month = oct,
	year = {2020},
	note = {ISSN: 2381-8549},
	pages = {2416--2420},
}

@misc{ramalho_density_2019,
	title = {Density estimation in representation space to predict model uncertainty},
	url = {http://arxiv.org/abs/1908.07235},
	urldate = {2023-09-27},
	publisher = {arXiv},
	author = {Ramalho, Tiago and Miranda, Miguel},
	month = oct,
	year = {2019},
	note = {arXiv:1908.07235 [cs, stat]},
}

@inproceedings{oberdiek_classification_2018,
author = {Oberdiek, Philipp and Rottmann, Matthias and Gottschalk, Hanno},
title = {Classification Uncertainty of Deep Neural Networks Based on Gradient Information},
year = {2018},
isbn = {978-3-319-99977-7},
publisher = {Springer-Verlag},
address = {Berlin, Heidelberg},
url = {https://doi.org/10.1007/978-3-319-99978-4_9},
doi = {10.1007/978-3-319-99978-4_9},
booktitle = {Artificial Neural Networks in Pattern Recognition: 8th IAPR TC3 Workshop, ANNPR 2018, Siena, Italy, September 19–21, 2018, Proceedings},
pages = {113–125},
numpages = {13},
location = {Siena, Italy}
}

@inproceedings{nandy_towards_2021,
author = {Nandy, Jay and Hsu, Wynne and Lee, Mong Li},
title = {Towards maximizing the representation gap between in-domain \& out-of-distribution examples},
year = {2020},
isbn = {9781713829546},
publisher = {Curran Associates Inc.},
address = {Red Hook, NY, USA},
booktitle = {Proceedings of the 34th International Conference on Neural Information Processing Systems},
articleno = {775},
numpages = {12},
location = {Vancouver, BC, Canada},
series = {NIPS '20}
}

@inproceedings{blundell_weight_2015,
author = {Blundell, Charles and Cornebise, Julien and Kavukcuoglu, Koray and Wierstra, Daan},
title = {Weight uncertainty in neural networks},
year = {2015},
publisher = {JMLR.org},
booktitle = {Proceedings of the 32nd International Conference on International Conference on Machine Learning - Volume 37},
pages = {1613–1622},
numpages = {10},
location = {Lille, France},
series = {ICML'15}
}

@inproceedings{hernandez-lobato_probabilistic_2015,
author = {Hern\'{a}ndez-Lobato, Jos\'{e} Miguel and Adams, Ryan P.},
title = {Probabilistic backpropagation for scalable learning of Bayesian neural networks},
year = {2015},
publisher = {JMLR.org},
booktitle = {Proceedings of the 32nd International Conference on International Conference on Machine Learning - Volume 37},
pages = {1861–1869},
numpages = {9},
location = {Lille, France},
series = {ICML'15}
}

@InProceedings{gal_dropout_2016,
  title = 	 {Dropout as a Bayesian Approximation: Representing Model Uncertainty in Deep Learning},
  author = 	 {Gal, Yarin and Ghahramani, Zoubin},
  booktitle = 	 {Proceedings of The 33rd International Conference on Machine Learning},
  pages = 	 {1050--1059},
  year = 	 {2016},
  editor = 	 {Balcan, Maria Florina and Weinberger, Kilian Q.},
  volume = 	 {48},
  series = 	 {Proceedings of Machine Learning Research},
  address = 	 {New York, New York, USA},
  month = 	 {20--22 Jun},
  publisher =    {PMLR},
  url = 	 {https://proceedings.mlr.press/v48/gal16.html},
}

@book{bishop_pattern_2009,
	address = {New York},
	edition = {13. (corrected at 8th printing 2009)},
	series = {Information science and statistics},
	title = {Pattern recognition and machine learning},
	isbn = {978-0-387-31073-2},
	language = {en},
	publisher = {Springer},
	author = {Bishop, Christopher M.},
	year = {2009},
}

@article{wilson_case_2020,
	title = {The {Case} for {Bayesian} {Deep} {Learning}},
	url = {http://arxiv.org/abs/2001.10995},
	urldate = {2021-02-09},
	journal = {arXiv:2001.10995 [cs, stat]},
	author = {Wilson, Andrew Gordon},
	month = jan,
	year = {2020},
	note = {arXiv: 2001.10995},
}

@inproceedings{louizos_multiplicative_2017,
author = {Louizos, Christos and Welling, Max},
title = {Multiplicative normalizing flows for variational Bayesian neural networks},
year = {2017},
publisher = {JMLR.org},
booktitle = {Proceedings of the 34th International Conference on Machine Learning - Volume 70},
pages = {2218–2227},
numpages = {10},
location = {Sydney, NSW, Australia},
series = {ICML'17}
}

@inproceedings{barber_ensemble_nodate,
  title={Ensemble learning in Bayesian neural networks},
  author={David Barber and Charles M. Bishop},
  year={1998},
  url={https://api.semanticscholar.org/CorpusID:14932413},
  booktitle = {Neural Networks and Machine Learning}
}

@article{wang_aleatoric_2019,
title = {Aleatoric uncertainty estimation with test-time augmentation for medical image segmentation with convolutional neural networks},
journal = {Neurocomputing},
volume = {338},
pages = {34-45},
year = {2019},
issn = {0925-2312},
doi = {https://doi.org/10.1016/j.neucom.2019.01.103},
url = {https://www.sciencedirect.com/science/article/pii/S0925231219301961},
author = {Guotai Wang and Wenqi Li and Michael Aertsen and Jan Deprest and Sébastien Ourselin and Tom Vercauteren},
}

@article{tuia_toward_2021,
	title = {Toward a {Collective} {Agenda} on {AI} for {Earth} {Science} {Data} {Analysis}},
	volume = {9},
	issn = {2168-6831, 2473-2397, 2373-7468},
	url = {https://ieeexplore.ieee.org/document/9456877/},
	doi = {10.1109/MGRS.2020.3043504},
	language = {en},
	number = {2},
	urldate = {2023-09-25},
	journal = {IEEE Geoscience and Remote Sensing Magazine},
	author = {Tuia, Devis and Roscher, Ribana and Wegner, Jan Dirk and Jacobs, Nathan and Zhu, Xiaoxiang and Camps-Valls, Gustau},
	month = jun,
	year = {2021},
	pages = {88--104},
}

@article{ayhan_test-time_2018,
	title={Test-time Data Augmentation for Estimation of Heteroscedastic Aleatoric Uncertainty in Deep Neural Networks},
	language = {en},
	author = {Ayhan, Murat Seçkin and Berens, Philipp},
    year = {2018},
    journal = {1st Conference on Medical Imaging with Deep Learning (MIDL)}
    
}

@inproceedings{lakshminarayanan_simple_2017,
author = {Lakshminarayanan, Balaji and Pritzel, Alexander and Blundell, Charles},
title = {Simple and scalable predictive uncertainty estimation using deep ensembles},
year = {2017},
isbn = {9781510860964},
publisher = {Curran Associates Inc.},
address = {Red Hook, NY, USA},
booktitle = {Proceedings of the 31st International Conference on Neural Information Processing Systems},
pages = {6405–6416},
numpages = {12},
location = {Long Beach, California, USA},
series = {NIPS'17}
}

@article{jospin_hands-bayesian_2022,
	title = {Hands-{On} {Bayesian} {Neural} {Networks}—{A} {Tutorial} for {Deep} {Learning} {Users}},
	volume = {17},
	issn = {1556-6048},
	url = {https://ieeexplore.ieee.org/document/9756596/},
	doi = {10.1109/MCI.2022.3155327},
	number = {2},
	urldate = {2025-07-24},
	journal = {IEEE Computational Intelligence Magazine},
	author = {Jospin, Laurent Valentin and Laga, Hamid and Boussaid, Farid and Buntine, Wray and Bennamoun, Mohammed},
	month = may,
	year = {2022},
	pages = {29--48},
}

@article{gawlikowski_survey_2021,
author = {Gawlikowski, Jakob and Tassi, Cedrique Rovile Njieutcheu and Ali, Mohsin and Lee, Jongseok and Humt, Matthias and Feng, Jianxiang and Kruspe, Anna and Triebel, Rudolph and Jung, Peter and Roscher, Ribana and Shahzad, Muhammad and Yang, Wen and Bamler, Richard and Zhu, Xiao Xiang},
title = {A survey of uncertainty in deep neural networks},
year = {2023},
issue_date = {Oct 2023},
publisher = {Kluwer Academic Publishers},
address = {USA},
volume = {56},
number = {Suppl 1},
issn = {0269-2821},
url = {https://doi.org/10.1007/s10462-023-10562-9},
doi = {10.1007/s10462-023-10562-9},
journal = {Artif. Intell. Rev.},
month = jul,
pages = {1513–1589},
numpages = {77}
}

@article{hullermeier_aleatoric_2021,
	title = {Aleatoric and epistemic uncertainty in machine learning: an introduction to concepts and methods},
	volume = {110},
	issn = {0885-6125, 1573-0565},
	shorttitle = {Aleatoric and epistemic uncertainty in machine learning},
	url = {http://link.springer.com/10.1007/s10994-021-05946-3},
	doi = {10.1007/s10994-021-05946-3},
	language = {en},
	number = {3},
	urldate = {2021-09-01},
	journal = {Machine Learning},
	author = {Hüllermeier, Eyke and Waegeman, Willem},
	month = mar,
	year = {2021},
	pages = {457--506},
}

@article{mukhoti_evaluating_2019,
	title = {Evaluating {Bayesian} {Deep} {Learning} {Methods} for {Semantic} {Segmentation}},
	url = {http://arxiv.org/abs/1811.12709},
	urldate = {2021-01-05},
	journal = {arXiv:1811.12709 [cs]},
	author = {Mukhoti, Jishnu and Gal, Yarin},
	month = mar,
	year = {2019},
	note = {arXiv: 1811.12709},
}

@misc{collier_simple_2020,
	title = {A {Simple} {Probabilistic} {Method} for {Deep} {Classification} under {Input}-{Dependent} {Label} {Noise}},
	url = {http://arxiv.org/abs/2003.06778},
	urldate = {2022-11-08},
	publisher = {arXiv},
	author = {Collier, Mark and Mustafa, Basil and Kokiopoulou, Efi and Jenatton, Rodolphe and Berent, Jesse},
	month = nov,
	year = {2020},
	note = {arXiv:2003.06778 [cs, stat]},
}

@inproceedings{kendall_what_2017,
author = {Kendall, Alex and Gal, Yarin},
title = {What uncertainties do we need in Bayesian deep learning for computer vision?},
year = {2017},
isbn = {9781510860964},
publisher = {Curran Associates Inc.},
address = {Red Hook, NY, USA},
booktitle = {Proceedings of the 31st International Conference on Neural Information Processing Systems},
pages = {5580–5590},
numpages = {11},
location = {Long Beach, California, USA},
series = {NIPS'17}
}

@misc{prapas_deep_2021,
	title = {Deep {Learning} {Methods} for {Daily} {Wildfire} {Danger} {Forecasting}},
	url = {http://arxiv.org/abs/2111.02736},
	urldate = {2023-02-20},
	publisher = {arXiv},
	author = {Prapas, Ioannis and Kondylatos, Spyros and Papoutsis, Ioannis and Camps-Valls, Gustau and Ronco, Michele and Fernández-Torres, Miguel-\'Angel and Guillem, Maria Piles and Carvalhais, Nuno},
	month = nov,
	year = {2021},
	note = {arXiv:2111.02736 [cs]},
}
